\documentclass{ieeeaccess}
\usepackage{amsmath,amssymb,amsfonts}
\usepackage{algorithmic}
\usepackage{graphicx}
\usepackage{textcomp}
\usepackage[hidelinks]{hyperref}
\def\BibTeX{{\rm B\kern-.05em{\sc i\kern-.025em b}\kern-.08em
    T\kern-.1667em\lower.7ex\hbox{E}\kern-.125emX}}

\def\etal{\textit{et al.}}
\usepackage{multirow}
\usepackage{comment}
\usepackage{subcaption}
\usepackage[numbers,sort]{natbib}
\usepackage{multirow}
\usepackage{array, booktabs, tabularx,makecell}
\usepackage{soul}
\newcommand{\sign}{\text{sign}}

\begin{document}
\history{Date of publication xxxx 00, 0000, date of current version xxxx 00, 0000.}
\doi{10.1109/ACCESS.2017.DOI}

\title{Adversarial Attacks and Defenses on 3D Point Cloud Classification: A Survey}
\author{\uppercase{Hanieh Naderi}\authorrefmark{1}, \IEEEmembership{Student Member,~IEEE},
\\
\uppercase{Ivan V. Baji\'{c}\authorrefmark{2}
\IEEEmembership{Senior Member,~IEEE},
}}
\address[1]{Department of Computer Engineering, Sharif University of Technology Tehran (e-mail: hanieh.naderii@gmail.com)}
\address[2]{School of Engineering
Science, Simon Fraser University, Burnaby, BC, Canada (e-mail: ibajic@ensc.sfu.ca)}

\begin{abstract}

Deep learning has successfully solved a wide range of tasks in 2D vision as a dominant AI technique. Recently, deep learning on 3D point clouds has become increasingly popular for addressing various tasks in this field. Despite remarkable achievements, deep learning algorithms are vulnerable to adversarial attacks. These attacks are imperceptible to the human eye, but can easily fool deep neural networks in the testing and deployment stage.
To encourage future research, this survey summarizes the current progress on adversarial attack and defense techniques on point-cloud classification. This paper first introduces the principles and characteristics of adversarial attacks and summarizes and analyzes adversarial example generation methods in recent years. Additionally, it provides an overview of defense strategies, organized into data-focused and model-focused methods.  Finally, it presents several current challenges and potential future research directions in this domain.

\end{abstract}


\titlepgskip=-15pt

\maketitle
\begingroup\renewcommand\thefootnote{\textsection}

\endgroup
\section{Introduction}
\label{sec:introduction}

\PARstart{D}{eep} learning (DL) \cite{lecun2015deep} is a subset of machine learning (ML) and artificial intelligence (AI) that analyzes large amounts of data 
using a structure roughly similar to the human brain. Deep learning is characterized by the use of multiple layers of neural networks, which process and analyze large amounts of data. These neural networks are trained on large datasets, which allows them to learn patterns and make decisions on their own. DL has achieved impressive results in the fields of image recognition \cite{li2022research,krizhevsky2017imagenet,naderi2020scale}, semantic analysis \cite{silberman2011indoor,mo2022review}, speech recognition \cite{nassif2019speech,taher2021deep} and natural language processing \cite{chowdhary2020natural} in recent years.

Despite the tremendous success of DL, in 2013 Szegedy \etal~\cite{szegedy2014intriguing} found that deep models are vulnerable to adversarial examples in image classification tasks. Adversarial examples are inputs to a deep learning model that have been modified in a way that is intended to mislead the model. In the context of image classification, for example, an adversarial example might be a picture of a panda that has been slightly modified in a way that is imperceptible to the human eye but that causes a deep learning model to classify the image as a gibbon. Adversarial examples can be created in two or three dimensions. In the case of 2D adversarial examples, the input is an image, and the modification is applied to the pixels of the image. These modifications can be small perturbations added to the image pixels~\cite{moosavi2016deepfool} or they can be more significant changes to the structure of the image~\cite{eykholt2018robust}.

Thanks to the rapid development of 3D acquisition technologies, various types of 3D scanners, LiDARs, and RGB-D cameras have become increasingly affordable. 3D data is often used as an input for Deep Neural Networks (DNNs) in 
healthcare~\cite{mozaffari2014systematic}, self-driving cars~\cite{badue2021self}, drones~\cite{hassanalian2017classifications}, robotics~\cite{pierson2017deep}, and many other applications. These 3D data, compared to 2D counterparts, 
capture more information from the environment, thereby allowing more sophisticated analysis. 
There are different representations of 3D data, like voxels~\cite{liu2019point}, meshes~\cite{ladicky2017point}, and point clouds~\cite{qi2017pointnet}. Since point clouds can be received directly from scanners, they can precisely capture shape details.
Therefore, it is the preferred representation for many safety-critical applications. Due to this, in the case of 3D adversarial examples, the input is a point cloud, and the modification is applied to the points in the cloud. These examples can be created by adding, dropping, and shifting some points in the input point clouds, or by generating entirely new point clouds with predefined target labels using methods such as Generative Adversarial Networks (GANs) or other transformation techniques.
It is typically easier to create adversarial examples in 2D space than in 3D space because the input space is smaller and there are fewer dimensions to perturb. 
In general, adversarial examples exploit the vulnerabilities or weaknesses in the model's prediction process, and they can be very difficult to detect because they are often indistinguishable from normal examples to the human eye. As a result, adversarial examples can pose a serious threat to the security and reliability of DL 
models. Therefore, it is important to have effective methods for defending against adversarial examples in order to ensure the robustness and reliability of DL 
models.

Adversarial defense in the 2D image and the 3D point clouds both seek to protect DL 
models from being fooled by adversarial examples. However, there are some key differences between the approaches used to defend against adversarial images and adversarial point clouds. Some of the main differences include the following:

\begin{itemize}
    \item Input data: Adversarial images are 2D data representations, while adversarial point clouds are 3D data representations. This means that the approaches used to defend against adversarial images and point clouds may need to take into account the different dimensions and characteristics of the input data.
    \item Adversarial perturbations: Adversarial images may be modified using small perturbations added to the image pixels, while adversarial point clouds may be modified using perturbations applied to individual points or groups of points in the point cloud. This means that the approaches used to defend against adversarial images and point clouds may need to be tailored to the specific types of adversarial perturbations that are being used.
    \item Complexity: Adversarial point clouds may be more complex to defend against than adversarial images, as the perturbations applied to point clouds may be more difficult to identify and remove. This may require the use of more sophisticated defenses, such as methods that are able to detect and remove adversarial perturbations from the input point cloud.
\end{itemize}

On the whole, adversarial point clouds can be challenging to identify and defend against, as they may not be easily recognizable in the 3D point cloud data. Adversarial point clouds may be more harmful and harder to defend against, because their changes may be less obvious to humans due to the lack of familiarity compared to images. 
As a result, it is important to conduct a thorough survey of adversarial attacks and defenses on 3D point clouds in order to identify the challenges and limitations of current approaches and to identify opportunities for future research in this area.
There are a number of published surveys that review adversarial attacks and defenses in general, including in the context of computer vision, ML, and AI systems. 
For example, Akhtar \etal~\cite{akhtar2018threat} focus on adversarial attacks in computer vision, with a particular emphasis on image and video recognition systems. Yuan \etal~\cite{yuan2019adversarial} delve into both adversarial attacks and defense mechanisms within the domain of images. Qiu \etal~\cite{qiu2019review} provide a comprehensive review of adversarial attacks and defenses in various AI domains, including image, video, and text. Wei \etal~\cite{wei2022physical} survey both attacks and defenses against physical 2D objects. Zhai \etal~\cite{zhai2023state} explore adversarial attacks and defenses within the context of graph-based data. Bountakas \etal~\cite{bountakas2023defense} review domain-agnostic defense strategies across multiple domains, including audio, cybersecurity, natural language processing (NLP), and computer vision. Pavlitska \etal~\cite{pavlitska2023adversarial} focus on adversarial attacks within the specific domain of traffic sign recognition, which is relevant to autonomous vehicles and road safety. These and several other surveys of adversarial attacks and defenses in various domains have been summarized in Table~\ref{surveys}. As seen in the table, there is a lack of surveys focused specifically on 3D point cloud attacks and defenses. Some published surveys do mention 3D attacks and defenses briefly, for example~\cite{akhtar2021advances}, but there is a need for more comprehensive surveys that delve deeper into this topic.



While our survey is focused on adversarial attacks and defenses on 3D point cloud classification, it is important to mention that there are existing general surveys on point cloud analysis and processing, which are not focused on adversarial attacks and defenses. For example, 
Guo \etal~\cite{guo2020deep} provide a comprehensive overview of deep learning methods for point cloud analysis, including classification, detection, and segmentation. Xiao \etal~\cite{xiao2023unsupervised} concentrate on unsupervised point cloud analysis. Nguyen \etal~\cite{xie2020linking} and Xie \etal~\cite{xie2020linking} specifically address point cloud segmentation tasks. Zhang \etal~\cite{zhang2023deep} focus on point cloud classification. Fernandes \etal~\cite{fernandes2021point} discuss point cloud processing in specialized tasks like self-driving, while Krawczyk \etal~\cite{krawczyk2023segmentation} tackle full human body geometry segmentation. Cao \etal~\cite{cao20193d} explore compression methods for 3D point clouds, essential for handling large data volumes. Although the focus of our survey is on 3D point cloud attacks and defenses, there is an intersection with some of the aforementioned surveys, especially in terms of models and datasets used for point cloud classification. We review the models and datasets that are relevant to the area of adversarial attacks and defenses, which can also be valuable resources for the broader community working on point cloud analysis and processing.

\begin{table*}
\begin{center}
\caption{A review of published surveys of adversarial attacks and defenses.}
\label{surveys} 
    \centering
    
    \begin{tabular}{ c  c  c  c }
    \toprule
    
    \bf Survey  &\bf Application Domain  &\bf Focus on &\bf Year  \\
    \midrule

    Akhtar \etal~\cite{akhtar2018threat} & Computer Vision (Image \& Video) & Attack & 2018  \\
    
    Yuan \etal~\cite{yuan2019adversarial} & Image & Attack \& defense &  2018 \\

    Qiu \etal~\cite{qiu2019review} & AI (Image \& Video \& Text) & Attack \& defense & 2019\\

    Wiyatno \etal~\cite{wiyatno2019adversarial} & ML (Image) & Attack & 2019\\

    Xu \etal~\cite{xu2020adversarial} & Image \& Graph \& Text & Attack \& defense & 2020 \\
    
    Martins \etal~\cite{martins2020adversarial} & Cybersecurity & Attack & 2020\\
        
    Chakraborty \etal~\cite{chakraborty2021survey} & Image \& Video & Attack \& defense & 2021 \\
    
    Rosenberg \etal~\cite{rosenberg2021adversarial} & Cybersecurity & Attack \& defense & 2021 \\    
    
    Akhtar \etal~\cite{akhtar2021advances} & Computer Vision (Image \& Video) & Attack \& defense  & 2021 \\
    
    Michel \etal~\cite{michel2022survey} & Image & Attack \& defense & 2022 \\
    
    Tan \etal~\cite{tan2022adversarial} & Audio & Attack \& defense & 2022 \\
    
    Qiu \etal~\cite{qiu2022adversarial} & Text & Attack \& defense & 2022 \\
    
    Liang \etal~\cite{liang2022adversarial} & Image & Attack \& defense & 2022 \\
    
    Li \etal~\cite{li2022review} & Image & Attack \& defense & 2022\\
    
    Gupta \etal~\cite{gupta2022adversarial} & AI (All) & Attack \& defense & 2022 \\

    Wei \etal~\cite{wei2022physically} & Physical 2D object & Attack \& defense & 2022 \\
    
      Wei \etal~\cite{wei2022physical} & Physical 2D object & Attack & 2022 \\
    
    Mi \etal~\cite{mi2022adversarial} & Object & Attack & 2022 \\

    Khamaiseh \etal~\cite{khamaiseh2022adversarial} & Image & Attack \& defense & 2022 \\

    Pavlitska \etal~\cite{pavlitska2023adversarial} & Image (Traffic Sign Recognition) & Attack & 2023 \\

    Kotyan \etal~\cite{kotyan2023reading} & ML & Attack & 2023 \\

    Zhai \etal~\cite{zhai2023state} & Graph & Attack \& defense & 2023 \\

   Baniecki \etal~\cite{baniecki2023adversarial} & AI (Image) & Attack \& defense & 2023 \\

    Han \etal~\cite{han2023interpreting} & Image & Attack & 2023 \\

    Bountakas \etal~\cite{bountakas2023defense} & Audio, cyber-security, NLP, \& computer vision & Defense & 2023 \\

    \bottomrule
    \end{tabular}

\end{center}
\end{table*}


Our key contributions are as follows:

\begin{itemize}
\item  A review of the different types of adversarial attacks on point clouds that have been proposed, including their methodologies and attributes, with specific examples from the literature. 
\item A review of the various methods that have been proposed for defending against adversarial attacks, organized into data-focused and model-focused methods, with examples from the literature. 
\item A summary of the most important datasets and models used by researchers in this field.
\item An overview 
of the challenges and limitations of the current approaches to adversarial attacks and defenses on 3D point clouds, and identification of opportunities for future research in this area.
\end{itemize}

\begin{figure*}[t]
    \centering
    \includegraphics[width=0.8\textwidth]{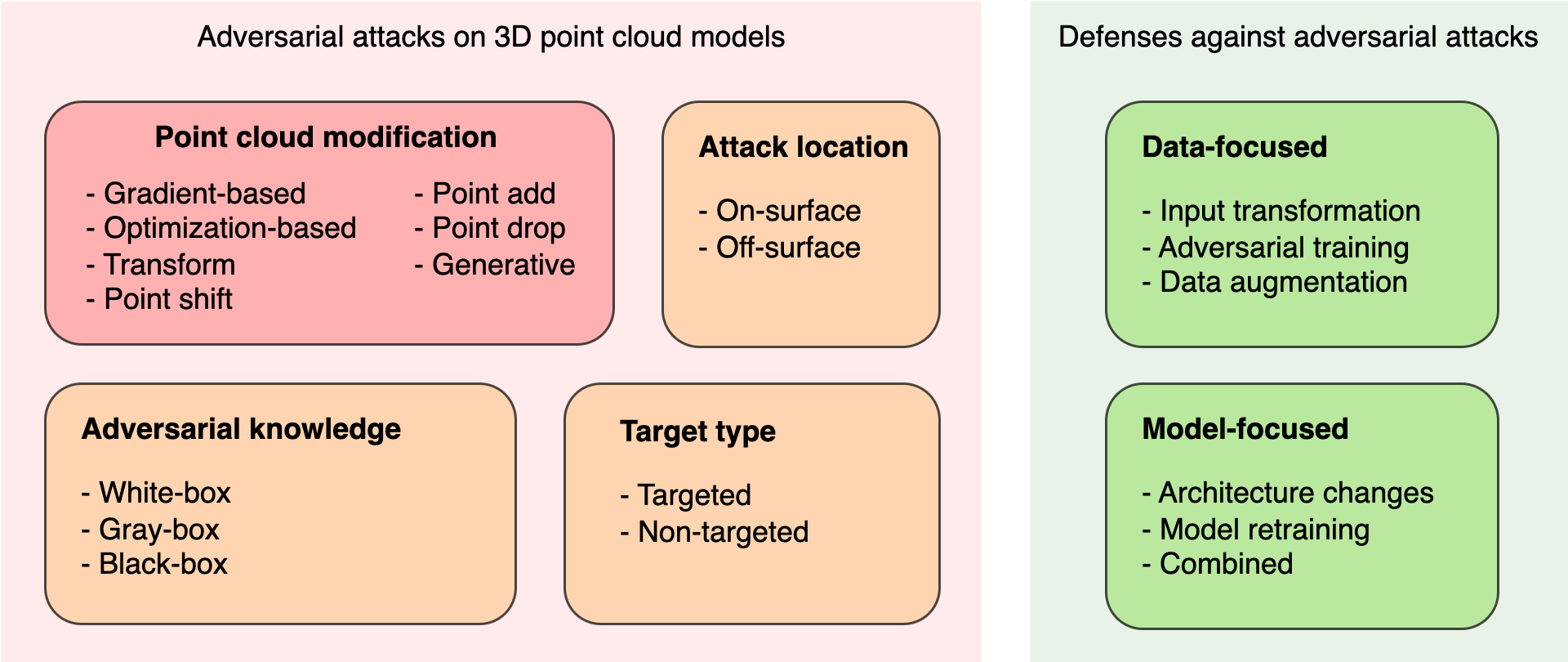}
    \caption{
    Categorization of adversarial attack and defense approaches on 3D point clouds.}
    \label{fig_overview}
\end{figure*}

An overview of the categorization of adversarial attack and defense approaches on 3D point clouds is shown in 
Fig.~\ref{fig_overview}. The rest of this paper is organized as follows. Section~\ref{sec:Backgrounds} introduces a list of notations, terms and measurements used in the paper. We discuss adversarial attacks on deep models for 3D point cloud classification 
in Section~\ref{sec:Adversarial attacks}. Section~\ref{sec:Defense} provides a detailed review of the existing adversarial defense methods. In Section~\ref{sec:Taxonomy}, we summarize commonly used datasets for point cloud classification and present an overview 
of datasets and victim models used in 
the area of adversarial attacks and defenses on point clouds. We discuss current challenges and potential future directions in Section~\ref{sec:Challenges}. Finally, Section~\ref{sec:Conclusion} concludes the survey.


\section{Background}
\label{sec:Backgrounds}
In this section, we provide the necessary background in terms of notation, terminology, and point cloud distance measures used in the field of 3D adversarial attacks. By establishing clear definitions, researchers can more accurately compare the effectiveness of different approaches and identify trends or patterns in the methods. 

A list of symbols used in the paper is given in Table~\ref{table:Symbols}, along with their explanations. These symbols are used to represent various quantities related to point cloud adversarial attacks. The table provides a brief description of each symbol to help readers understand and follow the discussions and 
equations in the paper. Next, we briefly introduce the terminology and distance measures used in the field of adversarial attacks and defenses on 3D point clouds.


\begin{table*}
\caption{
{Symbols and their explanations.}}
\centering
\label{table:Symbols}
\begin{tabular}{| c  | l |  }
\hline

\multicolumn{1}{|c|}{\bf Symbol}  & \multicolumn{1}{c|}{\bf Description}\\
\hline


$\mathcal{P}$ &  An instance of an original (input) point cloud \\

$\mathcal{P}^{adv}$ &  An instance of an adversarial point cloud \\



$p_i$ &  $i$-th point in the original (input) point cloud \\

$p^{adv}_i$ &  $i$-th point in the adversarial point cloud \\

$\eta$ &  Perturbation vector (difference between the original and adversarial point cloud) \\

$\epsilon$ & Perturbation threshold 
\\

$\alpha$ &   Scale parameter 
\\

$n$ &  Total number of points in a point cloud \\

$Y$ & ground-truth label associated with original input \\
$Y^{\prime}$ &  Wrong label associated with an adversarial example that deep model predicts \\
$T$ & Target attack label\\

$f(\cdot)$ & Mapping from the input point cloud to the output label implemented by the deep model  \\

$\theta$ & Parameters of model $f$  \\

$J(\cdot,\cdot)$ & Loss function used for model $f$  \\ 

$\nabla$ & Gradient \\

$\sign(\cdot)$ & Sign function \\

$P$ &  Parameter of the $\ell_P$-norm; typical values of $P$ are $1,2$ and $\infty$. \\

$\lambda$ &  Controls the trade-off between the two terms in the objective function \\


$D_{\ell_P}$ & $\ell_P$-norm distance \\
$D_{H}$ & Hausdorff distance \\

$D_{C}$ & Chamfer distance \\

 $k$ & Number of nearest neighbors of a point\\

 $\kappa$ & Confidence constant 
 \\

$z$ & Latent space of a point autoencoder\\
$g(\cdot)$ & Penalty function \\
$S(\cdot)$ & Statistical Outlier Removal (SOR) defense \\
$t$ & Number of iterations 
\\
$\mu$ & Mean of $k$ nearest neighbor distance of all points in a point cloud \\
$\sigma$ & Standard deviation of $k$ nearest neighbor distance of all points in a point cloud\\

\hline
\end{tabular}

\end{table*}


\subsection{Definition of terms}
\label{sec:Definition}

It is crucial to define the technical terms 
used in the literature in order to provide 
a consistent discussion of the various methods and approaches. The definitions of these terms appear below. The rest of the paper follows the same definitions 
throughout.

\begin{itemize}

\item \textbf{3D point cloud} is a set of 
points in 3D space, typically representing a 3D shape or scene.
\item \textbf{Adversarial point cloud} is a 3D point cloud that has been intentionally modified in order to mislead a DL model that analyzes 3D point clouds. We focus on geometric modifications, rather than attribute (e.g., color) modifications since these are predominant in the literature on adversarial point clouds.
\item \textbf{Adversarial attack} is a technique that intentionally introduces perturbations or noise to an input point cloud in order to fool a DL model, causing it to make incorrect predictions or decisions.
\item \textbf{Black-box attacks} are a type of adversarial attack in which the attacker only has access to the model's input and output and has no knowledge of the structure of the DL model being attacked.
\item \textbf{White-box attacks} are a type of adversarial attack in which the attacker knows all the details about the DL model’s architecture and parameters.
\item \textbf{Gray-box attacks} cover the spectrum between the extremes of black- and white-box attacks. Here, the attacker knows partial details about the DL model’s architecture and parameters in addition to having access to its input and output.
\item \textbf{Targeted attacks} involve manipulating the input point cloud in a way that causes the model to output 
a specific target label when presented with the modified input.
\item \textbf{Non-targeted attacks} involve manipulating the input point cloud in a way that causes the model to 
output a wrong label, regardless of what that label is.
\item \textbf{Point addition attacks} involve adding 
points to the point cloud 
to fool the DL model.
\item \textbf{Point shift attacks} involve shifting points of the point cloud to fool the DL model, while the number of points remains the same as in the original point cloud.
\item \textbf{Point drop attacks} involve dropping 
points 
from the 
point cloud to fool the DL model.
\item \textbf{Optimization-based attacks} are a type of attack in which 
the creation of an adversarial point cloud is formulated and solved as an optimization problem.
\item \textbf{Gradient-based attacks} are a type of attack in which 
the gradients of the loss function corresponding to each input point are 
used to generate an adversarial point cloud with a higher tendency toward being misclassified. 
\item \textbf{On-surface perturbation attacks} are a type of 
attack that involves modifying points along the object's surface in the point cloud.
\item \textbf{Off-surface perturbation attacks} are a type of 
attack that involves modifying points outside the object surface in the point cloud. 
\item \textbf{Transferability} refers to the ability of adversarial examples generated for one DL model to be successful in causing misclassification for another DL model. 
\item \textbf{Adversarial defense} is a set of techniques that aim to mitigate the impact of adversarial attacks and improve the robustness of the DL model against them.
\item \textbf{Attack success rate} refers to the percentage of times that an adversarial attack on a DL model is successful.
\end{itemize}

\subsection{Distance measures}
\label{sec:Measurement}

The objective of adversarial attacks is to modify the points of $\mathcal{P}$, creating an adversarial point cloud $\mathcal{P}^{adv}$, which could fool a DL model to produce wrong results. 
Geometric 3D adversarial attacks can be achieved by adding, dropping, or shifting points in $\mathcal{P}$. 
If the adversarial point cloud is generated by shifting points, $\boldsymbol{\ell_P}$\textbf{-norms} can be used to measure the distance between $\mathcal{P}$ and $\mathcal{P}^{adv}$, as the two 
point clouds have the same 
number of points. In this case, we can talk about the vector difference (perturbation) $\eta = \mathcal{P}-\mathcal{P}^{adv}$, and consider $\|\eta\|_P$ as the distance between $\mathcal{P}$ and $\mathcal{P}^{adv}$. The typical choices for $P$ are $P \in \{0, 2, \infty\}$, and the equation is:
\begin{equation}
 D_{\ell_P}  (\mathcal{P} , \mathcal{P}^{adv}) = \|\eta\|_P =  \left(\sum_{i=1}^n\|p_i - p^{adv}_i\|_P^P\right)^{1/P}
\label{eq:12}
\end{equation}
where $\mathcal{P} \in \mathbb{R}^{n\times3}$ is the original point cloud consisting of $n$ points in 3D space, $\mathcal{P}=\left\{p_{i}\,|\, i=1,2, ..., n\right\}$ and the $i^{th}$ point, $p_{i} = (x_i,y_i,z_i)$, is a 3D vector of coordinates. 
$\mathcal{P}^{adv}$ is the adversarial point cloud formed by adding the adversarial perturbation $\eta = (\eta_1,\eta_2, ..., \eta_n), \eta_i\in \mathbb{R}^3$, to $\mathcal{P}$. 
The three common $\ell_P$ norms have the following interpretations:

\begin{itemize}
\item $\boldsymbol{\ell_0}$\textbf{-norm} or $ \|\eta\|_0 $ counts the number of non-zero elements in $\eta$, so it indicates how many points in $\mathcal{P}^{adv}$ have changed compared to $\mathcal{P}$.
\item $\boldsymbol{\ell_2}$\textbf{-norm} or $ \|\eta\|_2 $ is the Euclidean distance between $\mathcal{P}^{adv}$ and $\mathcal{P}$.
\item $\boldsymbol{\ell_\infty}$\textbf{-norm} or $ \|\eta\|_\infty $ is the maximum difference between the points in $\mathcal{P}^{adv}$ and $\mathcal{P}$.
\end{itemize}


As mentioned above, $\ell_P$-norm distance criteria require that $\mathcal{P}^{adv}$ and $\mathcal{P}$ have the same number of points. Hence, these distance measures cannot be 
used for attacks
that involve adding or dropping points. To quantify the dissimilarity between two point clouds that don't have the same number of points, \textbf{Hausdorff distance} $D_H$ and \textbf{Chamfer distance} $D_C$ are commonly used. 
Hausdorff distance is defined as follows: 
\begin{equation}
D_{H} (\mathcal{P} , \mathcal{P}^{adv}) = \max\limits_{p \in \mathcal{P}} \min\limits_{p^{adv} \in \mathcal{P}^{adv}} \|p - p^{adv}\|_2^2
\label{eq:2}
\end{equation}
It locates the nearest original point $p$ for each adversarial point $p^{adv}$ and then finds the maximum squared Euclidean distance between all such nearest point pairs. Chamfer distance is similar to Hausdorff distance, except that it 
sums the distances among all pairs of closest points, instead of taking the maximum:
\begin{equation}
\begin{split}
D_{C} (\mathcal{P} , \mathcal{P}^{adv}) =& \sum_{p^{adv} \in \mathcal{P}^{adv}} \min\limits_{p \in \mathcal{P}}\|p - p^{adv}\|_2^2 \\
& + \sum_{p \in \mathcal{P}} \min\limits_{p^{adv} \in \mathcal{P}^{adv}}\|p - p^{adv}\|_2^2    
\end{split}
\label{eq:3}
\end{equation}
Optionally, Chamfer distance can be averaged with respect to the number of points in the two point clouds. 

In addition to the distance measures mentioned above, there are other distance measures for point clouds, such as the point-to-plane distance~\cite{PC_distortion_ICIP2017}, which are used in point cloud compression. However, these are not commonly encountered in the literature on 3D adversarial attacks, so we do not review them here.

\section{Adversarial attacks}
\label{sec:Adversarial attacks}

Various techniques have been proposed to generate adversarial attacks on 3D point cloud models. This section presents a classification of these attacks based on several criteria, illustrated in Fig.~\ref{fig_overview}. While different classifications are possible, ours is based on attack methodologies, such as gradient-based point cloud modification, etc., and attack attributes such as attack location (on-/off-surface), adversarial knowledge (white-box, gray-box, or black-box) and target type (targeted or non-targeted). 
In the following, we first present various methodologies, each with specific examples of the attack methods from that category. The discussion of various attack attributes is provided later in the section.
The most popular attack approaches are also summarized in Table~\ref{table:Categories} for quick reference. 


\begin{table*}
\centering
\caption{Popular adversarial attacks.}
\label{table:Categories}       
\begin{tabular}{c c|ccccccccccc}
\toprule
&  & \bf Targeted / & \bf Shift / Add / & \bf On- / & \bf Optimized / & \bf Black- / Gray- /\\ 

\bf Reference &  \bf Attack Name & \bf Non-targeted &  \bf Drop / Transform & \bf Off-surface &  \bf Gradient & \bf White-box \\
   


\midrule

\multirow{4}{*}{Xiang \etal~\cite{xiang2019generating}} & Perturbation  & Targeted & Shift & Off & Optimized & White\\& Independent points &Targeted & Add & Off & Optimized & White \\ & Clusters & Targeted & Add & Off &Optimized & White  \\ & Objects & Targeted & Add & Off &Optimized & White  \\
\hline
\multirow{2}{*}{Zheng \etal~\cite{zheng2019pointcloud}} & Drop100  & Non-Targeted & Drop & On& Gradient & White\\  & Drop200 & Non-Targeted & Drop & On& Gradient & White\\
\hline
\multirow{1}{*}{Hamdi \etal~\cite{hamdi2020advpc}} & Advpc & Targeted & Transform & On & Optimized & White\\
\hline
\multirow{1}{*}{Lee \etal~\cite{lee2020shapeadv}} & ShapeAdv & Targeted & Shift & On & Optimized & White\\
\hline
\multirow{1}{*}{Zhou \etal~\cite{zhou2020lg}} & LG-GAN & Targeted & Transform & On & - & White\\
\hline
\multirow{1}{*}{Wen \etal~\cite{wen2020geometry}} & GeoA$^3$ & Targeted & Shift & On & Optimized & White\\
\hline
\multirow{1}{*}{Tsai \etal~\cite{tsai2020robust}} & KNN & Targeted & Shift & On & Optimized & White\\
\hline
\multirow{1}{*}{Liu \etal~\cite{liu2019extending}} & Extended FGSM & Non-Targeted & Shift & Off & Gradient & White\\
\hline
\multirow{1}{*}{Arya \etal~\cite{arya2021adversarial}} & VSA & Non-Targeted & Add & On & Optimized & White\\
\hline
\multirow{4}{*}{Liu \etal~\cite{liu2020adversarial}} & Distributional attack & Non-Targeted & Shift & On & Gradient & White\\ & Perturbation resampling & Non-Targeted & Add & Off & Gradient & White\\ & Adversarial sticks & Non-Targeted & Add & Off & Gradient & White\\ & Adversarial sinks & Non-Targeted & Add & Off & Gradient & White\\
\hline
\multirow{2}{*}{Kim \etal~\cite{kim2021minimal}} & Minimal & Non-Targeted & Shift & Off & Optimized & White\\ & Minimal & Non-Targeted & Add & Off & Optimized & White\\
\hline
\multirow{1}{*}{Ma \etal~\cite{ma2020efficient}} & JGBA & Targeted & Shift & On & Optimized & White\\
\hline
\multirow{1}{*}{Liu \etal~\cite{liu2022imperceptible}} & ITA & Targeted & Shift & On & Optimized & Black\\
\hline
\multirow{4}{*}{Liu \etal~\cite{yang2019adversarial}} & FGSM & Non-Targeted & Shift & Off & Gradient & White\\& MPG  & Non-Targeted & Shift & Off & Gradient & White\\ & Point-attachment & Non-Targeted & Add & Off & Gradient & White\\& Point-detachment & Non-Targeted & Drop & On & Gradient & White\\
\hline
\multirow{1}{*}{Wicker \etal~\cite{wicker2019robustness}} & --- & Both & Drop & On & Optimized & Both\\

\hline
\multirow{1}{*}{He \etal~\cite{he2023generating}} & --- & Non-Targeted & Shift & On & Optimized & White\\

\bottomrule

\end{tabular}
\end{table*}

\subsection{Point cloud modification strategies}
\label{subsubsec: generation Strategies}

\subsubsection{Gradient-based strategies}
\label{subsubsec:Gradient-based}

DNNs are typically trained using the gradient descent method to minimize a specified loss function. Attackers targeting such models often take advantage of the fact that they can achieve their goals by maximizing this loss function along the gradient ascent direction. Specifically, attackers can create adversarial perturbations utilizing the gradient information of the model and iteratively adjusting the input to maximize the loss function. 
Fast Gradient Sign Method (FGSM) and Projected Gradient Descent (PGD) are the most commonly used gradient-based techniques for this purpose. We review each of them below.

\paragraph{3D Fast Gradient Sign Method (3D FGSM)}
\label{subsubsubsec:FGSM}

The inception of adversarial attacks on 3D data occurred in 2019 using gradient-based techniques. During this period, Liu \etal~\cite{liu2019extending} and Yang \etal~\cite{yang2019adversarial} extended the Fast Gradient Sign Method (FGSM), originally proposed by Goodfellow \etal~\cite{goodfellow2015explaining}, to 3D data. The 3D version of FGSM adds an adversarial perturbation $\eta$ to each point in the given point cloud $\mathcal{P}$ to create an adversarial point cloud  $\mathcal{P}^{adv} = \mathcal{P}+\eta$. Perturbations are generated according to the direction of the sign of the gradient at each point. 
The perturbation can be expressed as
\begin{equation}
\eta = \epsilon \cdot \sign\left[\nabla_\mathcal{P}J(f(\mathcal{P};\theta),Y)\right]
\label{eq:11}
\end{equation}
where $f$ is the deep model that is parameterized by $\theta$ and takes an input point cloud $\mathcal{P}$, and $Y$ denotes the label associated with $\mathcal{P}$. $J$ is the loss function,  $\nabla_\mathcal{P}J$ is its gradient with respect to $\mathcal{P}$ and $\sign(\cdot)$ denotes the sign function. The $\epsilon$ value is an adjustable hyperparameter that determines the $\ell_P$-norm of the difference between the original and adversarial inputs.



Liu \etal~\cite{liu2019extending} introduced three different ways  to define $\epsilon$ as a constraint for $\eta$ as follows


\begin{enumerate}
    \item Constraining the $\ell_2$-norm between each dimension of points in $\mathcal{P}$ and $\mathcal{P}^{adv}$.
    \item Constraining the $\ell_2$-norm between each point in $\mathcal{P}$ and its perturbed version in $\mathcal{P}^{adv}$.
    \item Constraining the $\ell_2$-norm between 
    the entire $\mathcal{P}$ and $\mathcal{P}^{adv}$.
\end{enumerate}
Due to the fact that the first method severely limits the movement of points, the authors suggest the second and third methods. However, all three methods have shown little difference in attack success rates.

\begin{figure*}
    \centering
    \includegraphics[width=\textwidth]{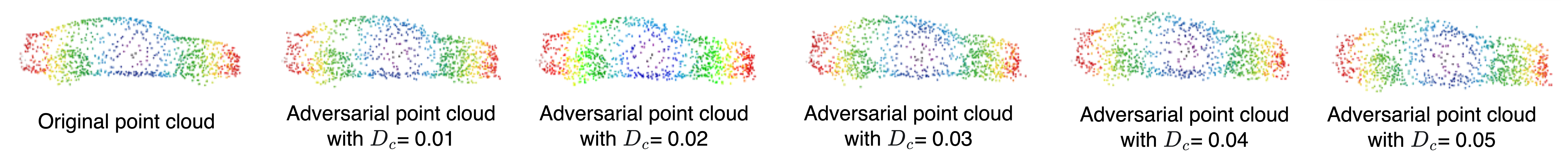}
    \caption{An example of original point cloud and 3D FGSM adversarial counterpart~\cite{yang2019adversarial} with Chamfer distances $D_c$ varying from 0.01 to 0.05. (Image source:~\cite{yang2019adversarial}; use permitted under the Creative Commons Attribution License CC BY 4.0.)}
    \label{fig_FGSM}
\end{figure*}

Yang \etal~\cite{yang2019adversarial}  used the Chamfer distance (instead of the $\ell_2$-norm) between the original point cloud and the adversarial counterpart to extend the FGSM to 3D. There is a trade-off between the Chamfer distance and the attack success rate because, as the Chamfer distance decreases, it may become more difficult for an adversarial attack to achieve a high attack success rate. However, if the Chamfer distance is set too high, the model may be more vulnerable to adversarial attacks. Finding the right balance between these two factors can be challenging, and it may depend on the specific characteristics of the point cloud model and the type of adversarial attack being used. Figure~\ref{fig_FGSM} illustrates an example of an FGSM adversarial point cloud with Chamfer distances varying from 0.01 to 0.05 between the two point clouds. The authors in~\cite{yang2019adversarial} set it to 0.02. 

Apart from the FGSM attack, Yang \etal~\cite{yang2019adversarial} introduced another attack called Momentum-Enhanced Pointwise Gradient (\textbf{MPG}). The (3D) MPG attack, similar to its 2D version~\cite{dong2018boosting}, integrates momentum into the iterative FGSM. The MPG attack produces more transferable adversarial examples because the integration of momentum into the iterative FGSM process enhances its ability to escape local minima and generate effective perturbations.

\paragraph{3D Projected Gradient Decent (3D PGD)}
\label{subsubsubsec:PGD}

One of the most potent attacks on 3D data is the Projected Gradient Descent (PGD), whose foundation is the pioneering work by Madry \etal~\cite{madry2019deep}. The iterative FGSM is considered a basis for PGD.
Taking the iterative FGSM method, we can generate the adversarial point cloud as
\begin{equation}
\begin{split}
    &\mathcal{P}^{adv}_{0} = \mathcal{P}, \\ &\mathcal{P}^{adv}_{t+1} = \text{clip}_{\mathcal{P},\epsilon}\left[\mathcal{P}^{adv}_t+\alpha \cdot \sign(\nabla_\mathcal{P}J(f(\mathcal{P};\theta),Y)\right],
\end{split}
\label{eq:10}
\end{equation}
where $t$ is the iteration number and $\text{clip}_{\mathcal{P},\epsilon}[\cdot]$ limits the change of the generated adversarial input to be within $\epsilon$ distance of $\mathcal{P}$, according to a chosen distance measure. 







The PGD attack is based on increasing the cost of the correct class $Y$, without specifying which of the incorrect classes the model should select. To do this, the PGD attack finds the perturbation $\eta$ that maximizes the loss function under the perturbation constraint controlled by $\epsilon$. The optimization problem can be formulated as:

\begin{equation}
\begin{split}
&\max\limits_\eta  \;\;\;    J(f(\mathcal{P}+\eta;\theta),Y)  \\
&\text{such that} \;\;\; D (\mathcal{P} , \mathcal{P}+\eta) \leq \epsilon
 \end{split}
\label{eq:50}
\end{equation}
where $J$ is the loss function and $\epsilon$ controls how far the adversarial point cloud can be from the original one according to the chosen distance measure $D$. 


Liu \etal~\cite{liu2020adversarial} proposed the following four flavors of the PGD attack.

\begin{figure}
    \centering
    \includegraphics[width=0.5\textwidth]{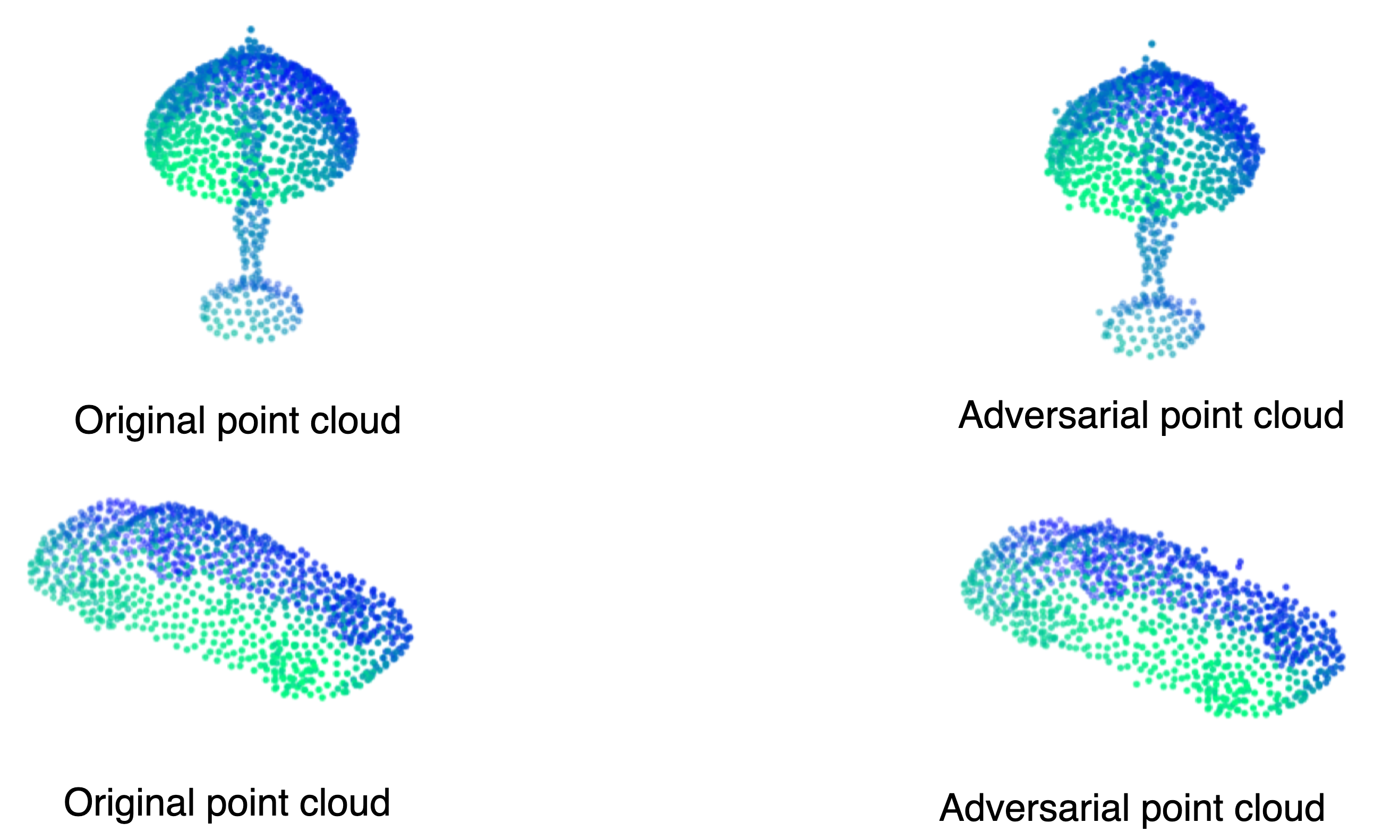}
    \caption{Two examples of the original point clouds (left) and adversarial point clouds generated by the distributional attack (right).~\cite{liu2020adversarial} (Image source:~\cite{liu2020adversarial}; use permitted under the Creative Commons Attribution License CC BY 4.0.).}
    \label{fig_PGD}
\end{figure}

\begin{enumerate}
\item \textbf{Perturbation resampling} This attack resamples a certain number of points with the lowest gradients by farthest point sampling to ensure that all points are distributed approximately uniformly. The algorithm is iterated to generate an adversarial point cloud that deceives the model. Hausdorff distance is used to maintain the similarity between $\mathcal{P}$ and $\mathcal{P}^{adv}$.

\item \textbf{Adding adversarial sticks} During this attack, the algorithm adds four sticks to the point cloud, such that one end is attached to the point cloud while the other end is a small distance away. The algorithm optimizes the two ends of the sticks so that the label of the point cloud is changed. 

\item \textbf{Adding adversarial sinks} In this case, critical points (the points remaining after max pooling in PointNet) are selected as ''sink'' points, which pull the other points towards them until the point cloud label is changed. The goal is to minimize global changes to non-critical points. The distance measure used to maintain the similarity between $\mathcal{P}$ and $\mathcal{P}^{adv}$ is $\ell_2$-norm.

\item \textbf{Distributional attack} This attack uses the Hausdorff distance between the adversarial point cloud and the triangular mesh fitted over the original point cloud, to push adversarial points towards the triangular mesh. This method is less sensitive to the density of points in $\mathcal{P}$ because it uses a mesh instead of the point cloud itself to measure the perturbation. 
Figure~\ref{fig_PGD} shows two examples of adversarial point clouds generated by the distributional attack.
\end{enumerate}


Ma \etal~\cite{ma2020efficient} proposed the Joint Gradient Based Attack (\textbf{JGBA}). They added an extra term to the objective function of the PGD attack~(\ref{eq:50}) to defeat statistical outlier removal (SOR), a common defense against attacks. 
Specifically, their optimization problem is:
\begin{equation}
\begin{split}
&\max\limits_\eta  \;\;\;    J(f(\mathcal{P}+\eta;\theta),Y) + \lambda \cdot J(f(S( \mathcal{P}+\eta);\theta),Y)   \\
&\text{such that} \;\;\; D_{\ell_2} (\mathcal{P} , \mathcal{P}+\eta) \leq \epsilon
 \end{split}
\label{eq:5000}
\end{equation}where $S(\cdot)$ denotes SOR and $\lambda$ is a hyperparameter that controls the trade-off between the two terms in the objective function. This way, the adversarial point cloud becomes more resistant against the SOR defense.

\begin{figure}
    \centering
    \includegraphics[width=0.4\textwidth]{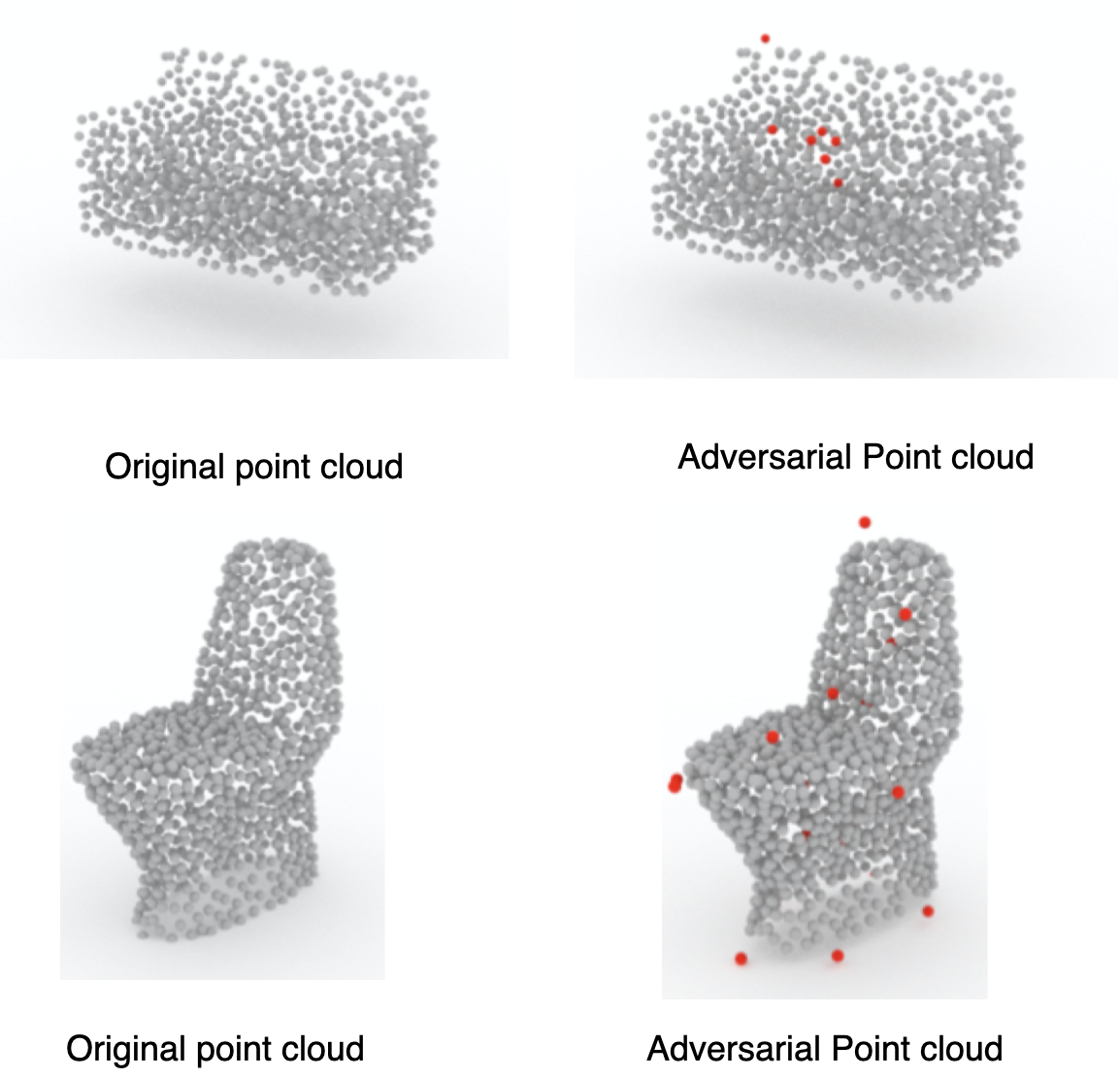}
    \caption{Two examples of the original point cloud and the corresponding minimal adversarial attack, where the altered points are shown in red~\cite{kim2021minimal} (© 2021 IEEE. Reprinted, with permission, from~\cite{kim2021minimal}).}
    \label{fig_min}
\end{figure}

Kim~\etal~\cite{kim2021minimal} proposed a so-called \textbf{minimal attack} that aims to manipulate a minimal number of points in a point cloud. This can be thought of as minimizing $D_{\ell_0}  (\mathcal{P} , \mathcal{P}^{adv})$. To find an adversarial point cloud, Kim~\etal~ modify the loss function of the PGD attack~(\ref{eq:30}) by adding a term that tries to keep the number of changed points to a minimum. Furthermore, they used Hausdorff and Chamfer distances to preserve the similarity between $\mathcal{P}$ and $\mathcal{P}^{adv}$. Figure~\ref{fig_min} illustrates examples of minimal adversarial attack, where the altered points are indicated in red.

\subsubsection{Optimization-based strategies}
\label{subsubsec:optimization-based}

While gradient-based strategies rely on model gradients, optimization-based methods instead utilize model output logits to create attacks. These methods usually aim to keep perturbations minimal, to reduce the chance of detecting the attack, while deceiving the model into making a wrong decision. Hence, these are often formulated as constrained optimization problems or multi-objective problems. 
The Carlini and Wagner (C\&W) attack is founded on these ideas, as explained below.

\paragraph{3D Carlini and Wagner attack (3D C\&W)}
\label{subsubsubsec:C&W}

The 3D version the C\&W attack was developed by Xiang \etal~\cite{xiang2019generating} as an extension of the original work by Carlini and Wagner~\cite{carlini2017towards}. The method can be described as an optimization problem of finding the minimum perturbation $\eta$ such that the output of the deep model to the adversarial input $\mathcal{P}^{adv}=\mathcal{P} + \eta$ is changed to the target label $T$. The problem can be formulated as 
\begin{equation}
\min\limits_\eta  \;\;\;    D (\mathcal{P} , \mathcal{P} + \eta) + c \cdot g(\mathcal{P} + \eta)  
\label{eq:30}
\end{equation}
where $D(\cdot,\cdot)$ is the distance measure, 
$c$ is a Lagrange multiplier and $g(\cdot)$ is a penalty function such that $g(\mathcal{P}^{adv})\leq$ 0 if and only if the output of the deep model is $f(\mathcal{P}^{adv})=T$. 
Seven choices for $g$ were listed in~\cite{carlini2017towards}. One of the functions evaluated in their experiments, which became popular in the subsequent literature, is as follows:
\begin{equation}
    g(\mathcal{P}^{adv}) = \max\left[\max\limits_{i=t}(Z(\mathcal{P}^{adv})_i)-Z(\mathcal{P}^{adv})_t , -\kappa\right]
\label{eq:4}
\end{equation}
where $Z$ denotes the Softmax function, and $\kappa$ represents a constant that controls confidence.
Compared to the FGSM attack, the C\&W attack does not constrain the perturbation; instead, it searches for the minimal perturbation that would produce the target label.

\begin{figure*}
    \centering
    \includegraphics[width=1\textwidth]{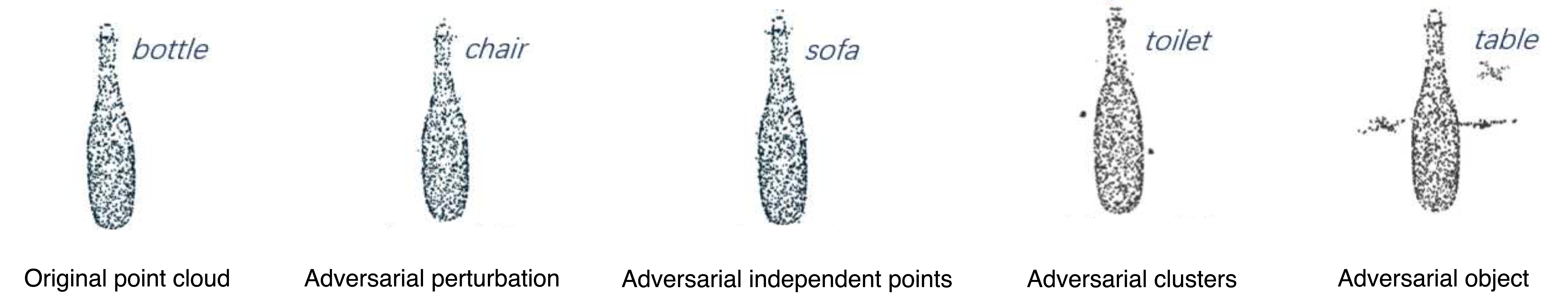}
    \caption{Left to right: original point cloud and the adversarial examples produced by the attacks proposed in~\cite{xiang2019generating} (© 2019 IEEE. Reprinted, with permission, from~\cite{xiang2019generating}).}
    \label{fig_CW}
\end{figure*}


Xiang \etal~\cite{xiang2019generating} developed four versions of the 3D C\&W attack, featuring various 
distance measures:

\begin{enumerate}

    \item \textbf{Adversarial perturbation} to shift the points toward the point cloud's surface, using the $\ell_2$-norm between all points of $\mathcal{P}$ and $\mathcal{P}^{adv}$ as the distance measure.
    
    \item \textbf{Adding adversarial independent points} by using two distance measures -- Hausdorff distance and Chamfer distance -- between $\mathcal{P}$ and $\mathcal{P}^{adv}$, to push independent points toward the point cloud's surface.

    \item \textbf{Adding adversarial clusters} based on three principles. 
    (1) Chamfer distance between the original point cloud and the adversarial cluster is used to push clusters toward the point cloud's surface.
    (2) Only a small number of clusters is added, specifically one to three. 
    (3) 
    The distance between the two most distant points in each cluster is minimized to constrain the added points clustered to be within small regions.
    
    \item \textbf{Adding adversarial objects} based on three principles.
    (1) Chamfer distance between the original point cloud and the adversarial object is used to push adversarial objects toward the point cloud's surface.
    (2) Only a small number of objects is added, specifically one to three. 
    (3) The $\ell_2$-norm between a real-world object and an adversarial object is used to generate shapes similar to those in the real world.
\end{enumerate}

Wen \etal~\cite{wen2020geometry} considered a new distance measure named \textit{consistency of local curvatures} to guide perturbed points towards object surfaces. Adopting the C\&W attack framework, the authors use a combination of the Chamfer distance, Hausdorff distance, and local curvature consistency as the distance measure to create a geometry-aware adversarial attack (\textbf{GeoA$^\mathbf{3}$}). The  GeoA$^3$ attack enforces the smoothness of the adversarial point cloud to make the difference between it and the original point cloud imperceptible to the human eye.
Finally, Zhang~\etal~\cite{zhang20233d} introduced a \textbf{Mesh Attack} designed to perturb 3D object meshes while minimizing perceptible changes. The Mesh Attack employs two key components in its loss function: a C\&W loss, encouraging misclassification of adversarial point clouds, and a set of mesh losses, including Chamfer, Laplacian, and Edge Length losses, to maintain the smoothness and geometric fidelity of the adversarial meshes relative to the original input mesh.


\subsubsection{Transform attacks}
\label{subsubsec:Transform attacks}

Transform attacks are crafted in the transform domain rather than the input domain. Usually, the 3D point cloud is transformed into another domain (e.g., 3D frequency domain), then modified, and then transferred back to the original input domain to be fed to the classifier. Liu~\etal~\cite{liu2022boosting} have suggested an adversarial attack based on the frequency domain, which aims to improve the transferability of generated adversarial examples to other classifiers.  They transformed the point cloud into the frequency domain using the graph Fourier transform (GFT)~\cite{GSP2013SPM}, then divided it into low- and high-frequency components, and applied perturbations to the low-frequency components to create an adversarial point cloud. 
Liu \etal~\cite{liu2022point} investigated the geometric structure of point clouds by perturbing, in turn, low-, mid-, and high-frequency components. They found that perturbing low-frequency components significantly changed their shape. To preserve the shape, they created an adversarial point cloud with constraints on the low-frequency perturbations and instead guided perturbations to the high-frequency components. Hu \etal~\cite{hu2022exploring} suggest that by analyzing the eigenvalues and eigenvectors of the graph Laplacian matrix~\cite{GSP2013SPM} of a point cloud, one can determine which areas of the cloud are particularly sensitive to perturbations. By focusing on these areas, the attack can be crafted more effectively.

A related attack, though not exactly in the frequency domain, was proposed by Huang \etal~\cite{huang2022shape}. This attack is based on applying reversible coordinate transformations to points in the original point cloud, which reduces one degree of freedom and limits their movement to the tangent plane. The best direction is calculated based on the gradients of the transformed point clouds. After that, all points are assigned a score to construct the sensitivity map. Finally, top-scoring points are 
moved to generate the adversarial point cloud.

In another attack called Variable Step-size Attack (\textbf{VSA})~\cite{arya2021adversarial}, a hard constraint on the number of modified points is incorporated into the optimization function of a PGD attack~(\ref{eq:30}) to try to preserve the point cloud's appearance. Specifically, points with the highest gradient norms, which are thought to have the greatest impact on classification, are selected initially. The selected points are then subject to adversarial perturbations. The goal is to shift these points in a way that maintains their original appearance while maximizing the loss function, thus causing the model to misclassify the input. By controlling the step size, VSA adjusts the magnitude of perturbations applied to the selected points. It starts with a larger step size to allow for rapid exploration of the optimization landscape. As the process advances, the step size is progressively reduced to guide the optimization toward more precise modifications.

\subsubsection{Point shift attacks}
\label{subsubsec:Point shift attacks}

Point shift attacks involve shifting the points of the original 3D point cloud to fool the deep model, while the number of points remains the same. Tsai \etal~\cite{tsai2020robust} developed a shifting point attack called K-Nearest Neighbor (\textbf{KNN}) attack that limits distances between adjacent points by adding another loss term to~(\ref{eq:30}). This additional loss term is based on the K-Nearest Neighbor distance for each point, while their main distance term in~(\ref{eq:30}) is the Chamfer distance. 
Miao \etal~\cite{zhao2020isometry} proposed an adversarial point cloud based on rotation by applying an isometry matrix to the original point cloud. To find an appropriate isometry matrix, the authors used the Thompson Sampling method \cite{russo2018tutorial}, which can quickly find a suitable isometry matrix with a high attack rate. 

Liu~\etal~\cite{liu2022imperceptible} proposed an Imperceptible Transfer Attack (\textbf{ITA}) that enhances the imperceptibility of adversarial point clouds by shifting each point in the direction of its normal vector. Although some of the attacks described earlier also involve shifting points, the main difference here is that ITA aims to make the attack imperceptible, whereas earlier attacks may cause noticeable changes to the shape. Along the same lines, Tang~\etal~\cite{tang2022normalattack} presented a method called \textbf{NormalAttack} for generating imperceptible point cloud attacks. Their method deforms objects along their normals by considering the object's curvature to make the modification less noticeable.

Zhao \etal~\cite{zhao2020nudge} proposed a class of point cloud perturbation attacks called Nudge attacks that try to minimize point perturbation while changing the classifier's decision. They generated adversarial point clouds using gradient-based and genetic algorithms with perturbations of up to 150 points to deceive the classifier. In some cases, the attack can fool the classifier by changing a single point when the point has a large distance from the surface of the objects.
Analogously to the one-pixel attack for images~\cite{su2019one}, Tan \etal~\cite{tan2023explainability} proposed an attack called \textbf{One point attack} in which only a single point in the point cloud needs to be shifted in order to fool the deep model. The authors also present a method to identify the most important points in the point cloud based on a saliency map, which could be used as candidates for the attack.

\subsubsection{Point add attacks}
\label{subsubsec:Point add attacks}

Point add attacks involve the addition of points to the point cloud with the aim of misleading deep models, while remaining plausible. Obviously, the number of points in the point cloud increases after this attack.
Yang~\etal~\cite{yang2019adversarial} provided a point-attachment attack by attaching a few points to the point cloud. Chamfer distance is used to keep the distance between the newly added points and the original point cloud small. Hard constraints limit the number of points added in the point cloud, making the adversarial point cloud preserve the appearance of the original one. 

Shape Prior Guided Attack~\cite{Shi2022Shape} is a method that adds points by using a shape prior, or prior knowledge of the structure of the object, to guide the generation of the perturbations. This method introduces Spatial Feature Aggregation (SPGA), which divides a point cloud into sub-groups and introduces structure sparsity to generate adversarial point sets. It employs a distortion function comprising Target Loss and Logical Structure Loss to guide the attack. The Shape Prior Guided Attack is optimized using the Fast Optimization for Attacking (FOFA) algorithm, which efficiently finds spatially sparse adversarial points. The goal of this method is to create adversarial point clouds that have minimal perturbations while still being able to fool the target classification model.

Note that some of the attack approaches described earlier also involve addition of points and can be considered to be point add attacks. For example, Liu \etal~\cite{liu2020adversarial} present several attacks such as \textbf{Perturbation resampling}, \textbf{Adding adversarial sticks} and \textbf{Adding adversarial sinks}, which can be considered point add attacks. These attacks were explained in more detail in Section~\ref{subsubsubsec:PGD}.

\subsubsection{Point drop attacks}
\label{subsubsec:Point drop attacks}

Attacks described in the previous sections 
involved adding or shifting points in the point domain or transform (latent) domain.  
This section reviews attacks that instead remove (drop) points from the point cloud to generate the adversarial point cloud. Obviously, the number of points in a point cloud reduces after a point drop attack. The points that are selected to be dropped are often referred to as ``critical'' points, in the sense that they are expected to be critical for a classifier to make the correct decision. Various methods have been developed to identify critical points in a point cloud.  

\begin{figure*}
    \centering
    \includegraphics[width=0.9\textwidth]{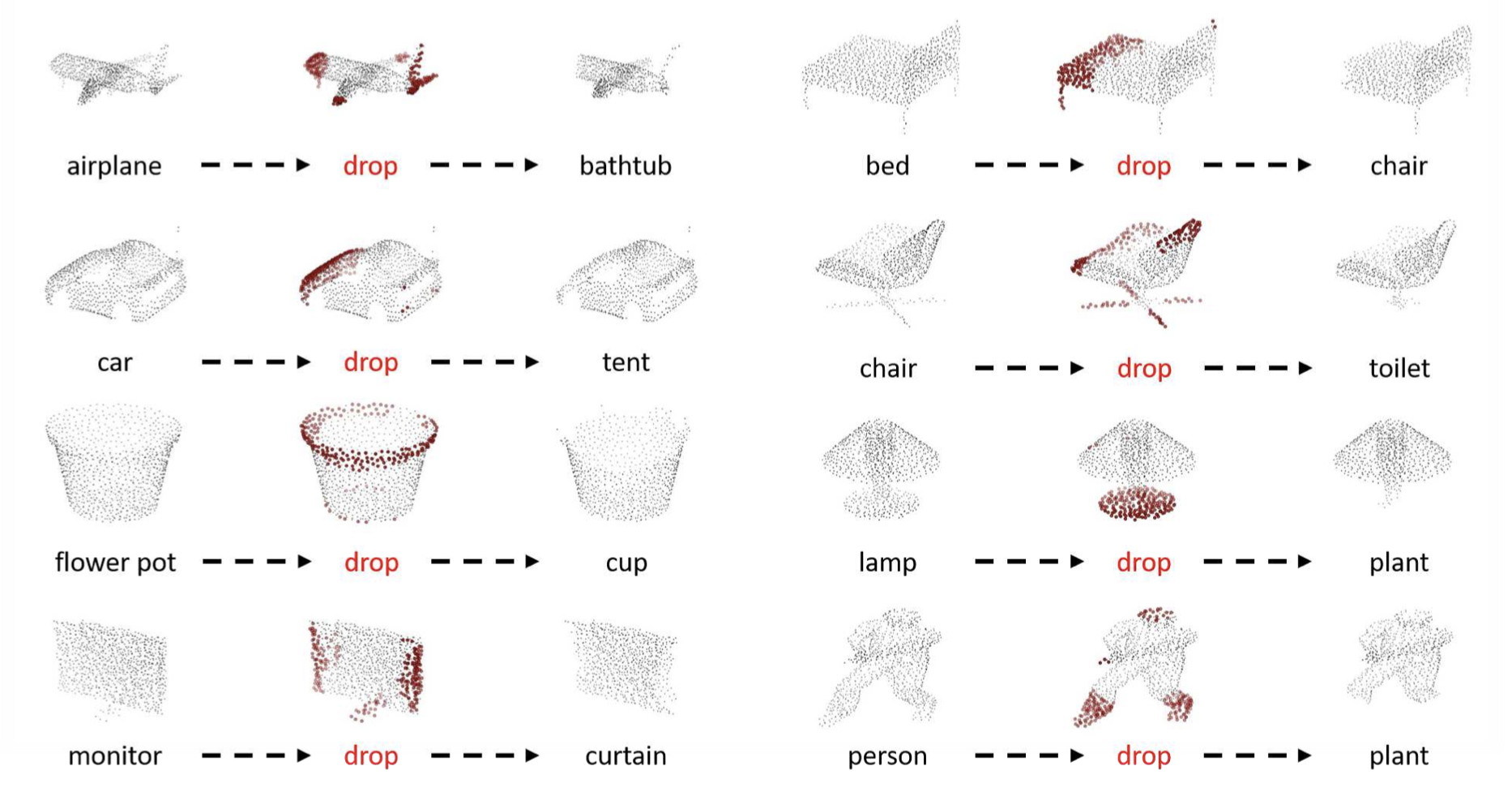}
    \caption{Original point clouds with labels (left), dropped points in red associated with highest scores (middle), and adversarial point clouds with new labels (right)~\cite{zheng2019pointcloud} (© 2019 IEEE. Reprinted, with permission, from~\cite{zheng2019pointcloud}).}
    \label{fig_saliency}
\end{figure*}

For example, Zheng \etal~\cite{zheng2019pointcloud} developed a method that uses a saliency map~\cite{zhou2016learning} to find critical points 
and drop them. As an illustration, critical points identified 
by high saliency values~\cite{zhou2016learning} are illustrated in red in Figure~\ref{fig_saliency}. The figure also shows what happens when these points are dropped. A version of this attack exists where, instead of dropping high-saliency points, they are shifted towards the point cloud center, thereby altering the shape of the point cloud in a manner similar to dropping points. 
Two versions of this attack have become popular in the literature, 
\textbf{Drop100} and \textbf{Drop200}, which drop 100 and 200 points, respectively.

An attack described in~\cite{naderi2022model} identifies ``adversarial drop points'' in a 3D point cloud that, when dropped, significantly reduce a classifier's accuracy. These points are specified by analyzing and combining fourteen-point cloud features, independently of a target classifier that is to be fooled. In this way, the attack becomes more transferable to different classifier models. 

In~\cite{wicker2019robustness}, the critical points are randomly selected and checked for dropping one by one. If dropping a point increases the probability of changing the ground-truth label  $f(\mathcal{P}) = Y$, such point is considered a critical point and will be dropped. Otherwise, it will not be dropped. This procedure continues iteratively until 
the classifier's decision is changed. The process can be described 
as the following optimization problem
\begin{equation}
\begin{split}
&\min\limits_{\mathcal{P}^{adv} \subseteq \mathcal{P}}   \;\;\;    (|\mathcal{P}| - |\mathcal{P}^{adv}|) \\
&\text{such that}  \;\;\;  f(\mathcal{P}^{adv}) \neq f(\mathcal{P})  
 \end{split}
\label{eq:60}
\end{equation}
where $|\mathcal{P}|$ and $|\mathcal{P}^{adv}|$ are the number of points in the original and adversarial point cloud, respectively. 

In order to determine the level of 
influence of a given point in PointNet decision-making, Yang \etal~\cite{yang2019adversarial} introduced a \textbf{Point-detachment attack} that assigned a \textit{class-dependent importance} to each point. A greedy strategy was employed to generate an adversarial point cloud, in which the most important points dependent on the ground-truth label are dropped iteratively. The class-dependent importance associated with a given point was determined by multiplying two terms. The first term used the PointNet feature matrix before max-pooling aggregation 
and the second term used the gradients relative to the ground-truth label output. The combination of these terms helped determine which points had the highest impact on the PointNet decision. 

\subsubsection{Generative strategies}
\label{subsubsec:generative-based}

Generative approaches utilize models such as Generative Adversarial Networks (GANs) and variational autoencoder models to create adversarial point clouds. Most of these attacks~\cite{zhou2020lg,lee2020shapeadv,tang2023deep,dai2021generating} attempt to change the shape of the point cloud in order to fool the deep model. The concept of these attacks can be related to what is called unrestricted attacks in 2D images~\cite{brown2018unrestricted,naderi2021generating,song2018constructing}. When such attacks occur, the input data might change significantly while remaining plausible. 
These attacks can fool the classifier without making humans confused. In this regard, Lee \etal~\cite{lee2020shapeadv} proposed shape-aware adversarial attacks called \textbf{ShapeAdv} that are based on injecting an adversarial perturbation $\eta$ into the latent space $z$ of a point cloud autoencoder. Specifically, the original point cloud is processed using an autoencoder to generate an adversarial point cloud, then the adversarial point cloud is fed to the classifier. Lee \etal~\cite{lee2020shapeadv} proposed three attacks with varying distance measures, which are used as a term in the C\&W loss to maintain similarity between the original and adversarial point clouds. All three attacks calculate the gradient of the C\&W loss with respect to the adversarial perturbation of the latent representation $z$. The three attacks are as follows: 
\begin{enumerate}
\item \textbf{Shape-aware attack in the latent space.} Here, the goal is to minimize the $\ell_2$-distance between the latent representation $z$ and the perturbed representation $z+\eta$. Using this approach, the original and adversarial point clouds are close in the latent space, but they could be highly dissimilar in the point space. 

\item \textbf{Shape-aware attack in the point space.} In this case, Chamfer distance is used to encourage the similarity of the original and adversarial point cloud in the point space. This is an attempt 
to resolve the issues with the previous attack, where the original and adversarial point cloud could be very different in the point space. 

\item \textbf{Shape-aware attack with auxiliary point clouds.} This attack minimizes the Chamfer distance between the adversarial point cloud and an auxiliary point cloud, which is created as the average of $k$ nearest neighbors sampled from the same class as the original point cloud. The goal is to avoid large adversarial perturbations in any direction in the latent space. To guide 
this process, point clouds sampled from the class of the original point cloud are used.
\end{enumerate}

\label{subsubsec:GAN}
Hamdi \etal~\cite{hamdi2020advpc} proposed an attack called \textbf{Advpc} by using an autoencoder that could be transferred between classification networks. The autoencoder was trained using a combination of two loss functions: the C\&W loss when the adversarial point cloud is fed directly to the classifier, and the C\&W loss when the point cloud is first fed to the autoencoder to project a perturbed point cloud onto the natural input manifold, then reconstructed, and then fed to the classifier. This strategy improved the transferability of the attack to different classification networks.

Tang \etal~\cite{tang2023deep} proposed a deep \textbf{manifold attack} that deforms the intrinsic 2-manifold structures of 3D point clouds. The attack strategy comprises two steps. In the first step, an autoencoder is used to establish a representation of the mapping between a 2D parameter plane and the underlying 2-manifold surface of the point cloud. This representation serves as a basis for subsequent transformations. The second step involves learning stretching operations within the 2D parameter plane. This stretching produces a 3D point cloud that can fool a pretrained classifier, all while keeping geometric distortion minimal.

\begin{figure}
    \centering
    \includegraphics[width=0.4\textwidth]{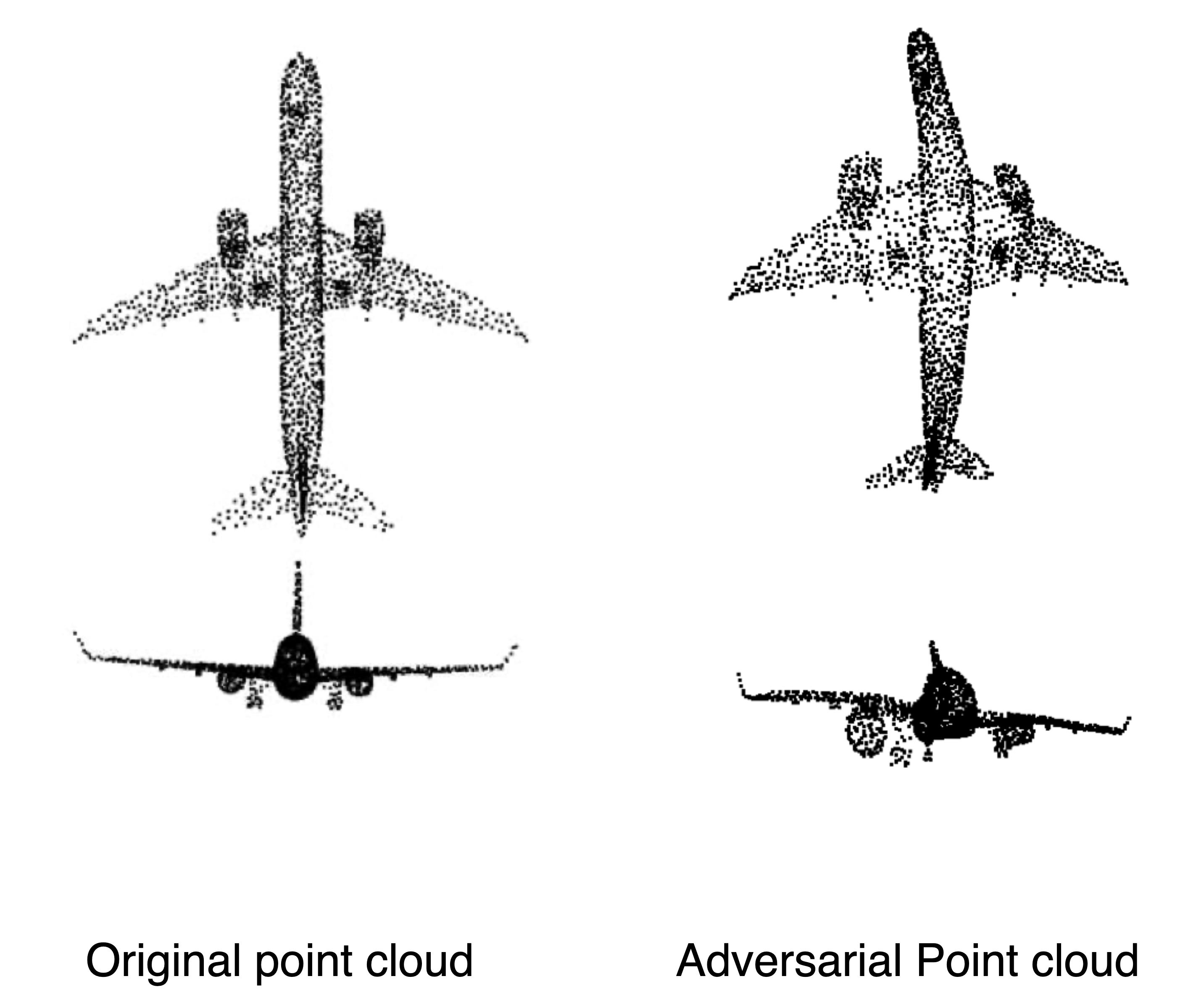}
    \caption{An example of original point cloud and LG-GAN attack were proposed in~\cite{zhou2020lg} (© 2020 IEEE. Reprinted, with permission, from~\cite{zhou2020lg}).}
    \label{fig_lggan}
\end{figure}

\textbf{LG-GAN} attack~\cite{zhou2020lg} generates an adversarial point cloud based on a Generative Adversarial Network (GAN). The GAN is trained using the original point clouds and target labels to learn how to generate adversarial point clouds to fool a classifier. It extracts hierarchical features from original point clouds, 
then integrates the specified label information into multiple intermediate features using the label encoder. The encoded features are fed into a reconstruction decoder to generate the adversarial point cloud. Once the GAN is trained, the attack is very fast because it only takes one forward pass to generate the adversarial point cloud. Figure~\ref{fig_lggan} shows an instance of the LG-GAN attack.

Dai \etal~\cite{dai2021generating} proposed another GAN-based attack, where the input to the GAN is noise, 
rather than the original point cloud. The noise vector and the target label are fed into a graph convolutional GAN, which outputs the generated adversarial point cloud. The GAN is trained using a 
four-part loss function including the objective loss, the discriminative loss, the outlier loss, and the uniform loss. 
The objective loss encourages the victim network to assign the (incorrect) target label to the adversarial point cloud while the discriminative loss encourages an auxiliary network to classify the adversarial point cloud correctly. The outlier loss and the uniform loss 
encourage the generator to preserve the point cloud shape.
Besides these GAN-based attacks, Lang \etal~\cite{lang2021geometric} proposed an attack that alters the reconstructed geometry of a 3D point cloud 
using an autoencoder trained on semantic shape classes, while Mariani \etal~\cite{mariani2020generating} proposed a method for creating adversarial attacks on surfaces embedded in 3D space, under weak smoothness assumptions on the perceptibility of the attack. 




\subsection{Attack location}
\label{subsubsec:perturbation location}

The location of perturbations plays a crucial role in changing the shape and distribution of points within a point cloud. This can result in points being shifted either off the object's surface, introducing noise, or along the surface, thereby altering the distribution of points.
Hence, in terms of the location of the perturbations, attacks can be categorized into two groups: on-surface and off-surface.

\textbf{On-surface perturbation attacks} are those attacks in which the points of the adversarial cloud $\mathcal{P}^{adv}$ are located along the object's original surface. Notably, drop attacks~\cite{zheng2019pointcloud,wicker2019robustness,yang2019adversarial} are an example of such attacks, since drop attacks involve solely the removal of points from the point cloud, so the remaining points stay on the original surface. While other attack methods like point shift or transform would normally tend to move the points off the object's surface, various approaches can be employed to keep the points at or near the original surface. For example, Hamdi~\etal~\cite{hamdi2020advpc} employ an autoencoder that projects off-surface perturbations onto the natural input manifold, thereby minimizing the movements of points off the surface. Another example is provided by Tsai~\etal~\cite{tsai2020robust}, who developed the KNN attack. This approach introduces constraints on the distances between adjacent points by adding an extra loss term based on the K-Nearest Neighbor distances for each point, with the goal of keeping the perturbations on the original surface. 
In the VSA attack~\cite{arya2021adversarial}, the magnitude of perturbations applied to adversarial points is adjusted by controlling the step size, which again could be used to keep points on the surface. The ``distributional attack''~\cite{liu2020adversarial} employs the Hausdorff distance between the adversarial point cloud and the triangular mesh fitted over the original point cloud. In this way, perturbed points can be guided toward the triangular mesh, effectively keeping the perturbed points at or near the object's original surface.

\textbf{Off-surface perturbation attacks} produce adversarial point clouds $\mathcal{P}^{adv}$ that include points off the original object's surface. As noted earlier, many strategies for generating adversarial point clouds include a distance term $D(\mathcal{P},\mathcal{P}^{adv})$ between the original and adversarial point cloud. This distance term is either involved in a hard constraint (upper bounded by an explicit value) or included as a loss term in the overall loss function. However, if its hard constraint is too high or if its scaling factor in the loss function is too low compared to other terms, this can result in off-surface points. For example, Yang \etal~\cite{yang2019adversarial} set the upper bound on the Chamfer distance to $0.2$, which is somewhat high and could result in off-surface perturbations. In other cases, off-surface perturbations are intentionally created. For example, Liu~\etal~\cite{liu2020adversarial}, in their adversarial sticks attack, add four sticks to the point cloud. These sticks are attached at one end to the point cloud and extend a small distance away, thereby creating an off-surface attack. Similarly, Xiang \etal~\cite{xiang2019generating} introduce clusters, individual points, or small objects off the surface of the object to craft their attacks. It should be noted that off-surface attacks might be easier to detect since the perturbed cloud's appearance starts deviating more obviously from the original one.

\subsection{Adversarial knowledge}
\label{subsubsec:Adversarial Knowledge}

In the context of adversarial knowledge, attacks can be categorized into three classes: white-box, black-box, and gray-box attacks. This classification is based on the extent of the attacker's knowledge about the target model. White-box and black-box scenarios represent extremes, whereas the gray-box scenario covers a wide range of possibilities between these extremes.

\textbf{White-box attacks} are those in which the attacker has complete information about the DL model under attack. This includes knowledge of the model's architecture, parameters, loss function, training details, and input/output training data. In the literature on adversarial attacks on 3D point cloud models, white-box attacks are quite common. Examples in this category include various gradient-based attack methods, such as those by Zhang \etal~\cite{zheng2019pointcloud} and Liu \etal~\cite{liu2020adversarial}. These approaches make use of the gradients of the loss function, propagated back through the model, to construct a variety of attacks such as point shifting, addition, and dropping. Other examples include attacks developed by Xiang \etal~\cite{xiang2019generating} and Liu \etal~\cite{liu2019extending,liu2020adversarial}, among others.

\textbf{Black-box attacks} are those in which the attacker has limited information -- sometimes none -- about the target model being attacked. In this case, at most, the attacker has access to the target model as a ``black box,'' meaning that it can generate the output of the model for a given input but lacks knowledge of the model's internal structure, training details, etc. Black-box attacks align more closely with real-world attack scenarios, but they are more difficult to construct. 

Black-box attacks are less common in the literature on adversarial attacks on 3D point cloud models. One example of a black-box attack is the ``model-free'' approach of Naderi \etal~\cite{naderi2022model}, which does not require any knowledge of the target model and instead focuses on identifying critical points within point clouds. This method takes advantage of the inherent properties of point clouds, bypassing the need for knowledge about the target model, and is therefore applicable to any model. Huang \etal~\cite{huang2022shape} proposed two versions of their attack, one white-box and the other black-box. The black-box attack relies on queries and saliency maps generated from a separate white-box surrogate model to craft adversarial perturbations that fool the target model.

Wicker and Kwiatkowska in~\cite{wicker2019robustness} randomly select and test critical points, dropping them if it increases the likelihood of changing the label. This iterative process continues until the model's decision changes. Therefore, the approach requires the ability to input a point cloud into the target model and access the output, but not the internal details of the target model, making it a black-box attack. ITA, by Liu \etal~\cite{liu2022imperceptible}, is another approach that could be classified as a black-box attack. It is based on subtly shifting each point in the point cloud along its normal vector, for which the knowledge of the internal architecture of the target model is not needed.  

The term \textbf{gray-box attack} has appeared in the literature more recently~\cite{Vivek_2018_ECCV}. It is intended to capture various scenarios between the extremes of white-box and black-box attacks. It should be noted that the boundaries of what are considered white- or black-box attacks are not crisp, and some variations in their interpretations do exist. For example, in a white-box scenario, the attacker may have access to all the internal parameters of the target model, but might not use all of them in constructing the attack. Therefore, such an attack can also be classified as a gray-box attack. That being said, so far very few approaches in the literature on 3D point cloud adversarial attacks have been declared gray-box, with notable exceptions in~\cite{dong2020self,Lang_3DV_2021}.

\subsection{Target type}
\label{subsubsec:Target type }

Some adversarial attacks attempt to guide the DL model towards a specific wrong label, while others simply want the model to produce any wrong label. The choice between these depends on the objectives of the attacker. Depending on the type of the label target, attacks can be classified as targeted or non-targeted.

\textbf{Targeted attacks} are those in which the goal is to make the DL model's output be a specific target label. There are two common approaches for choosing the target label in a targeted attack: 
\begin{enumerate}
    \item Most likely wrong label: Here, the target label is selected to be the one with the highest probability (confidence) other than the ground-truth label. The underlying idea is that this may be the easiest wrong label to lead the model towards.  
    \item Random label: Here, the target label is chosen randomly among the wrong labels. Although it might be harder to lead the model towards a randomly chosen label, such attack may be more impactful, especially in cases where the most likely wrong label is semantically close to the ground truth label (e.g., motorcycle vs. bicycle).
\end{enumerate}
The approach for target selection will depend on the specific objectives in a given scenario. For example, Wu \etal~\cite{wu2020if} and Naderi \etal~\cite{naderi2023lpf} have utilized the latter approach, while Ma \etal~\cite{ma2020efficient} have explored both strategies in their work.
 
\textbf{Non-targeted attacks} are those in which the goal is simply to make the DL model misclassify the input, regardless of which wrong label it eventually predicts. Examples of such attacks are those in~\cite{zheng2019pointcloud} and \cite{yang2019adversarial}, which operate by dropping points from the original point cloud until a label change occurs. Further examples of non-targeted attacks include ~\cite{hamdi2020advpc,lee2020shapeadv,zhou2020lg,wen2020geometry}. Interestingly, some studies present attacks that encompass both targeted and non-targeted types, for example~\cite{wicker2019robustness}. Here, a flexible attack framework is presented where, by encoding appropriate conditions and objectives through a Boolean function, both targeted and non-targeted attacks can be produced.

\section{Defenses against adversarial attacks}
\label{sec:Defense}

Adversarial defense methods for 3D point clouds can be data-focused or model-focused, as indicated in Fig.~\ref{fig_overview}. Data-focused strategies involve modifying the data on which the model is trained, or at inference time, in order to defend against attacks. Model-focused strategies may involve changing the model's architecture and/or retraining it to increase its robustness against attacks. Of course, combinations of these strategies are also possible. 
The following sections discuss defense methods under each of these categories. Moreover, we have provided an overview of the most prevalent defense strategies in Table~\ref{table:Categories_defenses} to simplify navigation and provide quick reference.


\begin{table*}
\centering
\caption{Categorization of defenses against adversarial attacks.}
\label{table:Categories_defenses}       
\begin{tabular}{c c | c c}
\toprule

\bf Reference &  \bf Defense Name & \bf Data- / Model-focused & \bf Type   \\
   


\midrule

\multirow{1}{*}{Yang \etal~\cite{yang2019adversarial}} & SRS & Data & Input transformation \\ 
\hline

\multirow{1}{*}{Zhou \etal~\cite{zhou2019dup}} & SOR & Data & Input transformation \\ 
\hline

\multirow{1}{*}{Liu \etal~\cite{liu2019extending}} & SRP & Data & Input transformation \\ 
\hline

\multirow{1}{*}{Zhou \etal~\cite{zhou2019dup}} & DUP-Net & Data & Input transformation \\ 
\hline

\multirow{1}{*}{Wu \etal~\cite{wu2020if}} & If-Defense & Data & Input transformation \\ 
\hline

\multirow{1}{*}{Liu \etal~\cite{liu2019extending}} & FGSM & Data & Adversarial training \\ 
\hline

\multirow{1}{*}{Liu \etal~\cite{liu2022imperceptible}} & ITA & Data & Adversarial training \\ 
\hline

\multirow{1}{*}{Liang \etal~\cite{liang2022pagn}} & PAGN & Data & Adversarial training \\ 
\hline

\multirow{1}{*}{Sun \etal~\cite{sun2021adversarially}} & --- & Data & Adversarial training \\ 
\hline

\multirow{1}{*}{Zhang \etal~\cite{zhang2022comprehensive}} & --- & Data & Data augmentation \\ 
\hline

\multirow{1}{*}{Yang \etal~\cite{yang2019adversarial}} & --- & Data & Data augmentation \\ 
\hline

\multirow{1}{*}{Zhang \etal~\cite{zhang2022pointcutmix}} & PointCutMix & Data & Data augmentation \\ 
\hline

\multirow{1}{*}{Naderi \etal~\cite{naderi2023lpf}} & LPF-Defense & Data & Data augmentation \\ 
\hline

\multirow{1}{*}{Zhang \etal~\cite{zhang2019defense}} & Defense-PointNet & Model & Deep model modification \\ 
\hline

\multirow{1}{*}{Zhang \etal~\cite{zhang2019defense}} & CCN & Model & Deep model modification \\ 
\hline

\multirow{1}{*}{Li \etal~\cite{li2022robust}} & LPC & Model & Deep model modification \\ 
\hline

\multirow{1}{*}{Sun \etal~\cite{sun2020adversarial}} & DeepSym & Combined & Deep model modification \& adversarial training \\ 

\bottomrule

\end{tabular}
\end{table*}


\subsection{Data-focused strategies}

\subsubsection{Input transformation}
\label{sec:Input transformation}

An input transformation is a preprocessing 
step that involves applying some transformation(s) to the input point cloud before it is fed into the deep model. These transformations could be designed to reduce the sensitivity of the deep model to adversarial attacks or to make it more difficult for an attacker to craft an adversarial point cloud. Input transformation methods are listed below.

\paragraph{Simple Random Sampling (SRS)}
\label{sec:SRS}
Simple random sampling (\textbf{SRS})~\cite{xiang2019generating} is a statistical technique that randomly drops a certain number of points (usually 500) from an input point cloud, with equal probability. It is crude but very fast. Many attacks involve shifting or adding points to a point cloud to cause a deep model to make an error. Random removal of points may remove some of these deliberately altered/inserted points and thereby make it less likely for the model to make an error. 
%

\paragraph{Statistical Outlier Removal (SOR)}
\label{sec:SOR}
Adversarial attacks that involve adding or shifting points usually result in outliers. 
Based on this observation, Zhou~\etal~\cite{zhou2019dup} proposed a defense based on statistical outlier removal (\textbf{SOR}). Specifically, their method  removed 
a point in an adversarial point cloud if the average distance of the point to its $k$ nearest neighbors was larger than 
$\mu + \sigma\cdot\alpha$, where $\mu$ is the mean and $\sigma$ is the standard deviation of the distance of the $k$ nearest neighbors to 
other points in the 
point cloud. Scaling factor $\alpha$ depends on $k$ and in \cite{zhou2019dup}, the authors used 
$\alpha = 1.1$ and $k=2$. 
A similar defense method was proposed in~\cite{zhou2018deflecting}. The Euclidean distance between each point and its $k$-nearest neighbors was used to detect outliers, and points with high average distances were discarded as outliers.

\paragraph{Salient Point Removal (SPR)}
\label{sec:Salient points removal}
Conceptually, salient point removal (\textbf{SPR}) is related to SOR, except that the outliers here are identified differently. 
For example, Liu \etal~\cite{liu2019extending} assumed that the adversarial points have fairly large gradient values. Based on this assumption, 
this method calculates the saliency of each point using the gradient of the output class of the model 
with respect to each point, and then removes the points with high saliency scores. 

\paragraph{Denoiser and Upsampler Network (DUP-Net)}
\label{sec:DUP-Net}
The \textbf{DUP-Net} defense approach consists of two steps. The first is a ``denoising'' step using SOR to remove outliers. 
This results in a point cloud with fewer points than the input cloud. The second step is upsampling using 
an upsampler network~\cite{yu2018pu} to produce a denser point cloud. These two steps are meant to undo typical attacks that generate outliers (either by shifting or adding points) in order to fool the deep model. By removing outliers and then bringing back the density, DUP-Net is meant to approximate the original point cloud. 

\paragraph{IF-Defense}
\label{sec:IF-Defense}
\textbf{IF-Defense}~\cite{wu2020if} is a preprocessing technique 
whose first step is SOR to remove outliers from the input point cloud.
In the next step, two losses are used to optimize the coordinates of the remaining points under geometry and distribution constraints.
The geometry-aware loss tries to push points towards the surface to improve smoothness. To estimate the surfaces of objects, the authors train a separate implicit function network~\cite{peng2020convolutional,mescheder2019occupancy}. 
Because the outputs of implicit functions are continuous, the predicted surface is locally smooth. This reduces the impact of the remaining outliers.
The second, distribution-aware loss, encourages the points to have a uniform distribution by maximizing the distance between each point and its $k$-nearest neighbors. Accordingly, IF-Defense produces smooth, uniformly sampled point clouds. 

\begin{figure*}
    \centering
    \includegraphics[width=0.7\textwidth]{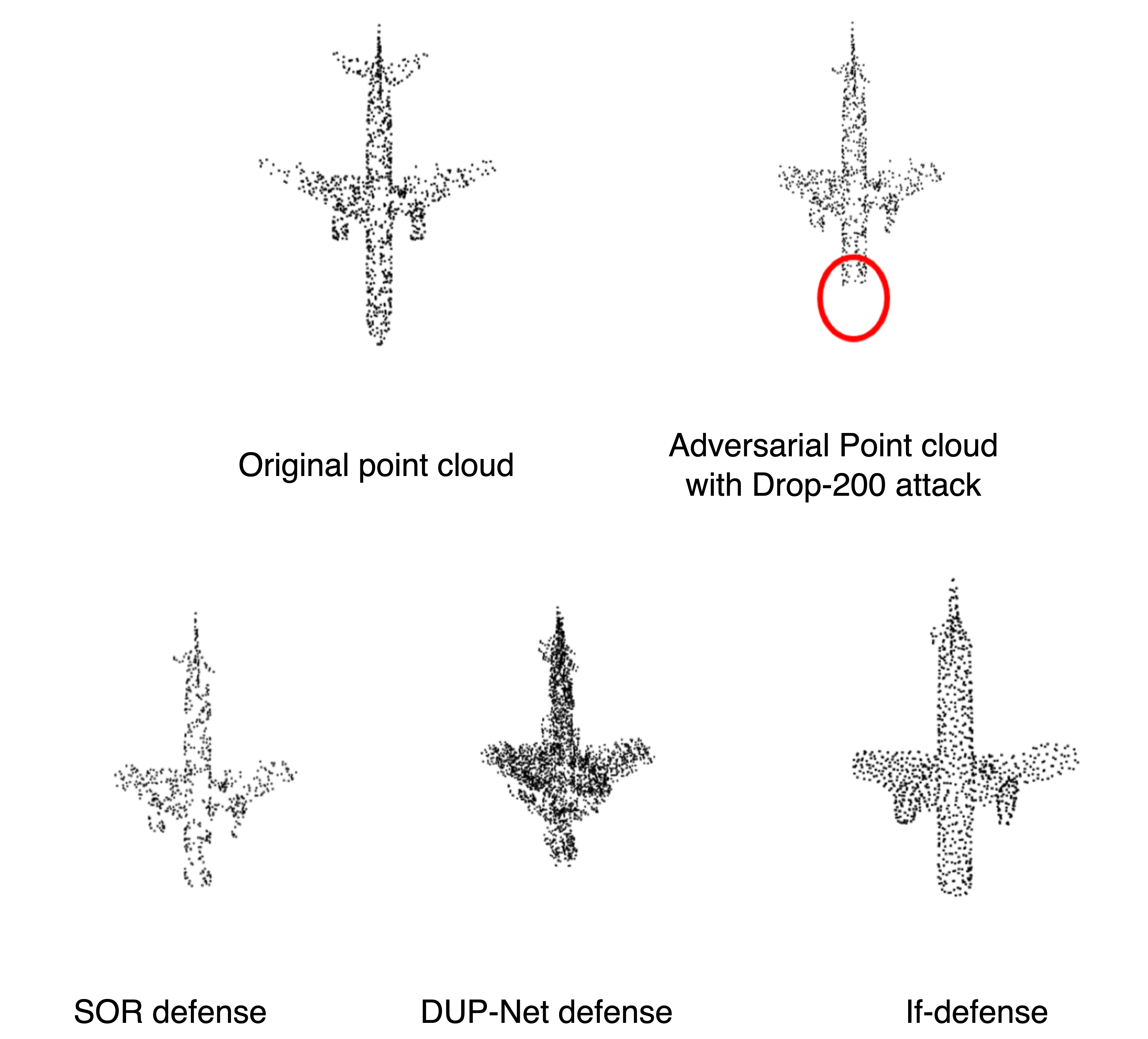}
    \caption{Results of three different defense methods applied on the Drop200 attack. Figure taken from~\cite{wu2020if} (Image source:~\cite{wu2020if}; use permitted under the Creative Commons Attribution License CC BY 4.0).}
    \label{fig_ifdefense}
\end{figure*}

Figure~\ref{fig_ifdefense} shows the results of three defense methods -- SOR, DUP-Net, and If-Defense -- against a Drop200 attack. As seen in the figure, SOR results in a relatively sparse point cloud, while DUP-Net produces a much denser cloud. IF-Defense produces a smooth, approximately uniformly sampled point cloud.

\paragraph{Miscellaneous defenses}
\label{sec:other Defenses}
Besides the above defenses, a few other approaches have been proposed to counter adversarial attacks through input transformation. Dong~\etal~\cite{dong2020self} proposed Gather-vector Guidance (\textbf{GvG}), which is sensitive to the change of local features. In case the adversarial perturbation changes the local features, the gather-vector will also change, thereby providing a way to detect the attack. 
Zhang \etal~\cite{zhang2022ada3diff} proposed \textbf{Ada3Diff}, which uses adaptive diffusion to smooth out perturbations in the point cloud. In doing so, it acts similarly to outlier removal, since the points perturbed during the attack often reduce local smoothness in order to fool the classifier. 

Liu~\etal~\cite{liu2021pointguard} developed an ensembling method called \textbf{PointGuard}. Here, a number of random subsets of the point cloud are taken and each is separately classified. Then 
the majority vote among the labels of these random subsets is taken as the final prediction. Similarly to SRS, the idea is that a random subset has fewer adversarially-perturbed points than the input point cloud, which may make it more likely to be classified correctly. An ensemble of such decisions makes the final prediction more robust.

\subsubsection{Training data optimization}
\label{sec:Data optimization}

Another group of defenses involves optimizing training data in order to make the trained model more robust against adversarial attacks. Various modifications to the training data have been proposed, as described below. 

\paragraph{Adversarial training}
\label{sec:Adversarial Training}
One way to make the model more robust against adversarial attacks is to expose it to adversarial examples during training, which is termed 
\textbf{adversarial training}~\cite{goodfellow2015explaining}. 
In adversarial training, both the original and adversarial point clouds are used. The use of adversarial training as a defense for point cloud models was first described in~\cite{liu2019extending}.  The authors of~\cite{liu2019extending} and~\cite{liu2022imperceptible} trained a deep model by augmenting the training data using adversarial examples generated by FGSM and ITA attacks. 
As a way to improve adversarial training, the authors of~\cite{liang2022pagn} employed adaptive attacks. Using this new adversarial training, different types of attacks are added to the deep model by embedding a perturbation-injection module. This module is utilized to generate the perturbed features for adversarial training. Sun~\etal~\cite{sun2021adversarially} applied self-supervised learning to adversarial training on point clouds.
 
\paragraph{PointCutMix}
\label{sec:PointCutMix}
Zhang~\etal~\cite{zhang2022pointcutmix} proposed the \textbf{PointCutMix} technique to generate a new training set by swapping points between two optimally aligned original point clouds and training a model on this new training set. 
PointCutMix provides two strategies for point swapping: randomly replacing all points or replacing the k nearest neighbors of a randomly chosen point. Additionally, the method uses a saliency map to guide point selection, enhancing its effectiveness. Augmented sample labels in PointCutMix are formed by blending the labels of the source point clouds. The augmented point clouds, along with their associated labels, are integrated into the training set, thereby creating a novel collection of training samples that capture variations from both original point clouds. Overall, PointCutMix proves valuable for augmenting point cloud data in tasks such as classification and defense against adversarial attacks.

\paragraph{Low Pass Frequency-Defense (LPF-Defense)}
\label{sec:LPF-Defense}
In LPF-Defense~\cite{naderi2023lpf}, deep models are trained with the low-frequency version of the original point clouds.  More specifically, using the Spherical Harmonic Transform (SHT)~\cite{cohen2018spherical}, original point clouds are transformed from the spatial to the frequency domain. Then the high-frequency components are removed and the low-frequency version of the point cloud is 
recovered in the spatial domain. 
The idea is that adversarial attacks, through point shifting, insertion, or deletion, often introduce high frequencies into the point cloud. When a deep model is trained on the low-frequency versions of the point clouds, it learns to associate the label with low frequencies and thereby implicitly ignores high frequencies which may have been introduced during an attack.

\subsection{Model-focused strategies}

\subsubsection{Deep model modification}
\label{sec:Deep model modification}

Another class of defenses involves 
modifying the architecture of the deep model itself and may involve retraining in order to improve its robustness to adversarial attacks. 
Examples of this type of defense are given below.

\paragraph{Defense-PointNet}
\label{sec:Defense-PointNet}
Zhang~\etal~\cite{zhang2019defense} proposed a defense method that involves splitting the PointNet model into two parts. The first part is the feature extractor, with a discriminator attached to its last layer. 
The second part is the remainder of the PointNet model, which acts as a classifier. 
The feature extractor is fed with 
a mini-batch consisting of the original point clouds and adversarial examples generated by the FGSM attack. 
The discriminator attempts to classify whether the features come from the original or adversarial point cloud.
Model parameters are optimized using three different loss functions: one for the classifier, one for the discriminator, and one for the feature extractor. While discriminator loss encourages the model to distinguish the original point cloud from the adversarial one, the feature extractor loss tries to mislead the discriminator to label every feature 
vector as the original. Therefore, the feature extractor acts as an adversary to the discriminator. Finally, the classifier loss encourages the classifier to give correct predictions for each input.

\paragraph{Context-Consistency dynamic graph Network (CCN)}
\label{sec:Context-Consistency dynamic graph Network}
Li \etal~\cite{li2022improving} proposed two methodologies to improve the adversarial robustness of 3D point cloud classification models. 
The first one involves a novel point cloud architecture named the Context-Consistency dynamic graph Network (CCN). This architecture is predominantly constructed upon the Dynamic Graph CNN (DGCNN) model~\cite{wang2019dynamic}, but it incorporates a lightweight Context-Consistency Module (CCM) into various layers of DGCNN. This module aims to reduce feature gaps between clean and noisy samples. 
The second one is a new data augmentation technique. In each training epoch, the method generates three types of batches from each sample: adversarial examples created by dropping points, adversarial examples created by shifting points, and clean samples. Subsequently, it dynamically identifies the most appropriate samples based on their accuracy to train the model, thereby adaptively balancing the model's accuracy and robustness to attacks.
To provide a more robust model against adversarial point clouds, the authors integrate the two methodologies.

\paragraph{Lattice Point Classifier (LPC)}
\label{sec:Lattice Point Classifier (LPC)}
Li~\etal~\cite{li2022robust} proposed embedding a declarative node into the networks to transform adversarial point clouds such that they may be classified more easily. 
Specifically, structured sparse coding in the permutohedral lattice~\cite{kiefel2015permutohedral} is used to construct a 
Lattice Point Classifier (LPC). The LPC projects each point cloud onto a lattice and generates a 2D image, which is then input to a 2D CNN for classification. 
Projection onto a lattice may 
remove some of the noise and/or outliers introduced during an adversarial attack.

\subsubsection{Model retraining}
Model (re)training strategies include adversarial (re)training, discussed in Section~\ref{sec:Adversarial Training}, but also other strategies intended to make the model more robust without explicitly using adversarial examples. Such strategies may involve various data augmentation methods, additional regularization terms to encourage robustness and generalization, as well as contrastive learning to robustify class boundaries. The authors of~\cite{zhang2022comprehensive,yang2019adversarial} augmented the training data by noise to make the resulting model more robust against attacks. One noise model employed was additive Gaussian noise, which was meant to improve robustness against point shifts in an attack. Another type of noise used was quantization noise, which involved converting point cloud coordinates to low precision during training. Quantization noise is often modeled as uniform noise~\cite{Widrow2008}, so this augmentation was meant to improve robustness against small point movements in a limited range.

\subsubsection{Combined strategies}
Some of the adversarial defense methods combine various strategies described above. 
For example, Sun \etal~\cite{sun2020adversarial} studied the role of pooling operations in enhancing model robustness during adversarial training. They found that fixed operations like max-pooling weaken the effectiveness of adversarial training, while sorting-based parametric pooling operations improve the model's robustness. As a result, they proposed \textbf{DeepSym}, a symmetric pooling operation that increases model's robustness to attacks.

\section{Datasets and victim models}
\label{sec:Taxonomy}

A variety of 3D point cloud datasets have been collected to train and evaluate deep models on point cloud classification. These include ModelNet~\cite{wu20153d}, ShapeNet~\cite{chang2015shapenet}, ScanObjectNN~\cite{uy2019revisiting}, McGill Benchmark~\cite{siddiqi2008retrieving}, ScanNet~\cite{dai2017scannet}, Sydney Urban Objects~\cite{de2013unsupervised}. 
A summary of these datasets and their unique characteristics is presented in Table~\ref{table:datasets}.

These 3D point cloud datasets can be broadly categorized into two groups: synthetic and real. ShapeNet and ModelNet are well-known datasets that contain synthetic data. These datasets are often used for model training and evaluation in controlled settings, because objects in synthetic datasets are typically complete, without occlusions, ``holes,'' and free of noise. For instance, ModelNet10 and ModelNet40 consist of 3D models of various objects, categorized into 10 and 40 classes, respectively, and are widely used in point cloud research. ShapeNet is a larger dataset with a larger number of classes, making it suitable for more challenging classification tasks. Virtual KITTI~\cite{gaidon2016virtual} is an example of a synthetic dataset built for autonomous driving.  

In contrast, datasets such as ScanNet and ScanObjectNN contain real data collected from real-world measurements, reflecting the complexity and variability of actual environments. ScanObjectNN is a real-world dataset suitable for evaluating 3D object classification in real-world scenarios. ScanNet is another real-world dataset that includes 3D scans of indoor environments.  KITTI~\cite{geiger2012we} is a real-world dataset featuring 3D scenes related to autonomous driving. Real 3D point-cloud scans are often subject to occlusion and may contain noise, which may necessitate ``hole filling''~\cite{dinesh2018inpainting} and/or denoising~\cite{dinesh2020point} before further use.
Among the datasets discussed above,  ModelNet10~\cite{wu20153d}, ModelNet40~\cite{wu20153d}, ShapeNet~\cite{chang2015shapenet} and ScanObjectNN~\cite{uy2019revisiting} have been very popular in the literature on point cloud adversarial attacks and defenses. 

Table~\ref{tab_Bechmark} presents an overview of prominent victim models that researchers commonly employ to assess adversarial attacks and defense strategies in the context of point cloud classification. PointNet, PointNet++, and DGCNN are the models that are the most frequently targeted for adversarial assessment.
Each of these models employs distinct mechanisms for processing point clouds. PointNet employs multi-layer perceptrons (MLPs) to extract pointwise features and aggregate them using max-pooling. PointNet++ builds upon PointNet, incorporating three key layers: the sampling layer, the grouping layer, and the PointNet-based learning layer. This architecture is repeated to capture fine geometric structures in point clouds.
DGCNN, another widely used model, leverages local geometric structures by constructing a local neighborhood graph and applying convolution-like operations on the edges connecting neighboring points.

Beyond these popular models, there are other notable architectures like PointConv, which extends the Monte Carlo approximation of 3D continuous convolution operators. It employs MLPs to approximate weight functions for each convolutional filter and applies density scaling to re-weight these learned functions.
The Relation-Shape Convolutional Neural Network (RS-CNN) extends regular grid CNNs to handle irregular point-cloud configurations. It achieves this by emphasizing the importance of learning geometric relations among points, forcing the convolutional weights to capture these relations based on predefined geometric priors.
VoxNet, on the other hand, is an architecture that combines a volumetric grid and a 3D CNN to improve object recognition using point cloud data from sensors like LiDAR and RGBD cameras. VoxNet predicts object class labels directly from the volumetric occupancy information. 

SpiderCNN is specifically designed for extracting geometric features from point clouds. It achieves this by using a family of convolutional filters parametrized as a combination of a step function, capturing local geodesic information, and a Taylor polynomial to enhance expressiveness.
PointASNL is capable of handling noisy point clouds effectively. Its core feature is the adaptive sampling module, which re-weights and adjusts sampled points to improve feature learning and mitigate the impact of outliers. It also includes a local-nonlocal module to capture local and global dependencies.
CurveNet addresses the limitations of existing local feature aggregation approaches by grouping sequences of connected points (curves) through guided walks in point clouds and then integrating these curve features with point-wise features.
Lastly, AtlasNet introduces a novel approach to 3D shape generation that does not rely on voxelized or point-cloud representations. Instead, it directly learns surface representations by deforming a set of learnable parameterizations.

\begin{table*}
\caption{
{ Summary of the datasets commonly used 3D point cloud classification.
}}
\centering
\begin{tabular}{ c  c  c  c  c }
\toprule
\bf Dataset & \bf Year & \bf Type & \bf Classes & \bf Samples (Training / Test) \\
\midrule
ModelNet10~\cite{wu20153d} & 2015 & Synthetic & 10 & 4899 (3991 / 605)\\

ModelNet40~\cite{wu20153d} & 2015 & Synthetic & 40 & 12311 (9843 / 2468)\\

ShapeNet~\cite{chang2015shapenet} & 2015 & Synthetic & 55 & 51190 ( / ) \\

ScanObjectNN~\cite{uy2019revisiting} & 2019 & Real & 15 & 2902 (2321 / 581)\\

KITTI~\cite{geiger2012we} & 2012 & Real & 8 & 7058 (6347 / 711) \\

Virtual KITTI~\cite{gaidon2016virtual} & 2016 & Synthetic & 8 & 21260 ( / ) \\

ScanNet~\cite{dai2017scannet} &2017 & Real & 17 & 12283 (9677 / 2606) \\

3DMNIST~\cite{3Dminst} & 2019 & Synthetic & 10 & 12000 (10000 / 2000) \\

McGill Benchmark~\cite{siddiqi2008retrieving} & 2008 & Synthetic & 19 & 456 (304 / 152) \\

Sydney Urban Objects~\cite{de2013unsupervised} & 2013 & Real & 14 & 588 ( / ) \\
\bottomrule
\end{tabular}
\label{table:datasets}
\end{table*}

\begin{table*}
\begin{center}
\caption{Summary of datasets and victim models used in attacks and defenses on 3D point clouds.}
\label{tab_Bechmark} 
\begin{tabular}{ c|c|c} 
\hline

\multirow{9}{*}{Datasets}
& \multirow{1}{*}{ModelNet10~\cite{wu20153d}} & \cite{wicker2019robustness}, \cite{sun2020adversarial}, \cite{sun2021improving}, \cite{zhao2020nudge}\\
\cline{2-3}
& \multirow{3}{*}{ModelNet40~\cite{wu20153d}} & \cite{liu2019extending}, \cite{wicker2019robustness}, \cite{liu2022imperceptible}, \cite{kim2021minimal}, \cite{liu2020adversarial}, \cite{arya2021adversarial}, \cite{tsai2020robust}, \cite{zhou2020lg}, \cite{wen2020geometry}, \cite{zheng2019pointcloud}\\ && \cite{xiang2019generating}, \cite{wu2020if}, \cite{huang2022shape}, \cite{tang2022rethinking}, \cite{liang2022pagn}, \cite{denipitiyage2021provable}, \cite{sun2020adversarial}, \cite{sun2021local}, \cite{sun2021improving}, \cite{zhang20233d}\\ && \cite{lee2020shapeadv}, \cite{dong2020self}, \cite{zhao2020nudge}, \cite{zhao2020isometry}, \cite{ma2020efficient}, \cite{zhou2019dup}, \cite{ma2021towards}, \cite{li2022robust}, \cite{miao2022isometric}, \cite{liu2022point}, \cite{Shi2022Shape}, \cite{he2023point}\\
\cline{2-3}
&\multirow{1}{*}{ShapeNet~\cite{chang2015shapenet}} & \cite{arya2021adversarial}, \cite{zhou2020lg}, \cite{hamdi2020advpc}, \cite{wu2020if}, \cite{tang2022rethinking}, \cite{zhao2020isometry}, \cite{lang2021geometric}, \cite{zhang2019defense}\\
\cline{2-3}
& \multirow{1}{*}{ScanObjectNN~\cite{uy2019revisiting}} & \cite{kim2021minimal}, \cite{denipitiyage2021provable}, \cite{liu2021pointguard}, \cite{sun2020adversarial}, \cite{sun2021improving} \\
\cline{2-3}

& \multirow{1}{*}{KITTI~\cite{geiger2012we}} & \cite{wicker2019robustness}, \cite{cheng2021universal} \\
\cline{2-3}

& \multirow{1}{*}{ScanNet~\cite{qi2017pointnet}} & \cite{liu2021pointguard}, \cite{li2022robust}  \\
\cline{2-3}

& \multirow{1}{*}{3D-MNIST~\cite{3Dminst}} & \cite{zheng2019pointcloud}, \cite{Shi2022Shape}\\
\hline

\multirow{18}{*}{Victim models}

& \multirow{3}{*}{PointNet~\cite{qi2017pointnet}} & \cite{liu2019extending}, \cite{wicker2019robustness}, \cite{yang2019adversarial}, \cite{liu2022imperceptible}, \cite{kim2021minimal}, \cite{liu2020adversarial}, \cite{arya2021adversarial}, \cite{zhou2020lg}, \cite{wen2020geometry}, \cite{hamdi2020advpc}\\
&& \cite{zheng2019pointcloud}, \cite{xiang2019generating}, \cite{wu2020if}, \cite{liang2022pagn}, \cite{liu2022boosting}, \cite{denipitiyage2021provable}, \cite{liu2021pointguard}, \cite{sun2020adversarial}, \cite{sun2021local}, \cite{sun2021improving}\\ && \cite{dai2021generating}, \cite{zhang20233d}, \cite{lee2020shapeadv}, \cite{dong2020self}, \cite{zhao2020nudge}, \cite{zhao2020isometry}, \cite{lang2021geometric}, \cite{zhang2019defense}, \cite{ma2020efficient}, \cite{zhou2019dup}\\ && \cite{cheng2021universal}, \cite{ma2021towards}, \cite{li2022robust}, \cite{miao2022isometric}, \cite{liu2022point}, \cite{Shi2022Shape}, \cite{he2023point}\\
\cline{2-3}

& \multirow{3}{*}{PointNet++~\cite{qi2017pointnet1}} & \cite{liu2019extending}, \cite{yang2019adversarial}, \cite{liu2022imperceptible}, \cite{kim2021minimal}, \cite{liu2020adversarial}, \cite{arya2021adversarial}, \cite{tsai2020robust}, \cite{zhou2020lg}, \cite{wen2020geometry}, \cite{hamdi2020advpc}\\ && \cite{zheng2019pointcloud}, \cite{xiang2019generating}, \cite{wu2020if}, \cite{huang2022shape}, \cite{tang2022rethinking}, \cite{liang2022pagn}, \cite{liu2022boosting}, \cite{sun2021local}, \cite{dai2021generating}, \cite{zhang20233d}\\ && \cite{lee2020shapeadv}, \cite{dong2020self}, \cite{zhao2020isometry}, \cite{ma2020efficient}, \cite{zhou2019dup}, \cite{cheng2021universal}, \cite{miao2022isometric}, \cite{liu2022point}, \cite{Shi2022Shape}, \cite{he2023point}\\
\cline{2-3}
& \multirow{3}{*}{DGCNN~\cite{phan2018dgcnn}} & \cite{yang2019adversarial}, \cite{liu2022imperceptible}, \cite{kim2021minimal}, \cite{liu2020adversarial}, \cite{arya2021adversarial}, \cite{zhou2020lg}, \cite{wen2020geometry}, \cite{hamdi2020advpc}, \cite{zheng2019pointcloud}, \cite{xiang2019generating}\\ && \cite{wu2020if}, \cite{huang2022shape}, \cite{tang2022rethinking}, \cite{liang2022pagn}, \cite{liu2022boosting}, \cite{liu2021pointguard}, \cite{sun2021improving}, \cite{dai2021generating}, \cite{zhang20233d}, \cite{lee2020shapeadv}\\ && \cite{zhao2020nudge}, \cite{zhao2020isometry}, \cite{ma2020efficient}, \cite{ma2021towards}, \cite{miao2022isometric}, \cite{liu2022point}, \cite{Shi2022Shape}, \cite{he2023point}\\
\cline{2-3}
&PointConv~\cite{wu2019pointconv} & \cite{wu2020if}, \cite{tang2022rethinking}, \cite{liu2022boosting}\\
\cline{2-3}
&RS-CNN~\cite{liu2019relation} & \cite{wu2020if}, \cite{he2023point}\\
\cline{2-3}
&VoxNet~\cite{maturana2015voxnet} & \cite{wicker2019robustness}   \\
\cline{2-3}
&SpiderCNN~\cite{xu2018spidercnn} & \cite{kim2021minimal}\\
\cline{2-3}
&PointASNL~\cite{yan2020pointasnl} & \cite{kim2021minimal}\\
\cline{2-3}
&CurveNet~\cite{xiang2021walk} & \cite{huang2022shape}\\
\cline{2-3}
&AtlasNet~\cite{groueix2018papier} & \cite{lang2021geometric}\\
\cline{2-3}
&PointTrans~\cite{zhao2021point} & \cite{liu2022point}\\
\cline{2-3}
&PointMLP~\cite{ma2022rethinking} & \cite{liu2022point}\\

\hline

\end{tabular}
\end{center}
\end{table*}

\section{Challenges and future directions}
\label{sec:Challenges}

In this section, we explore the current challenges within the domain of adversarial attacks and defenses on 3D point clouds. We also present several promising directions for future research in this area. 

\subsection{Current challenges}

\subsubsection{Crafting real-world attacks}
As mentioned earlier, majority of attacks on 3D point clouds reported in the literature are white-box attacks. However, in practice, the white-box scenario is much less likely compared to the black-box and gray-box scenarios. Existing results suggest that black-box attacks are much less effective than white-box attacks. Hence, one of the current challenges is developing attack strategies that do not rely on complete knowledge of the target model and whose effectiveness could approach that of white-box attacks. 


\subsubsection{Understanding the role of frequency} 

Points in a point cloud are irregularly placed in the 3D space. This makes understanding the frequency content of point clouds more challenging than that in the case of images or other regularly-sampled signals. Tools from graph signal processing~\cite{GSP2013SPM} or spherical harmonic analysis~\cite{cohen2018spherical} are useful in this context, but the fact remains that even the basic notion of frequency and its role in attacks and defenses is harder to analyze in the case of point clouds.
Many attacks introduce high frequencies into the point cloud through methods like point shifting or adding. But if the original point cloud already contains high frequencies, they may mask the attack and therefore make defenses less effective.

A better understanding of the role of frequency may help explain the reasons behind the vulnerability of 3D deep models to adversarial attacks. 
For example,~\cite{naderi2023lpf} has tackled this problem, suggesting that 3D deep models may rely too heavily on high-frequency details within 3D point clouds, and removing these details could potentially lead to models that are more robust against attacks. Such deeper understanding may be useful in the context of adversarial attacks and defenses in other areas, not just 3D point clouds. 


\subsubsection{Training for robustness}

As mentioned earlier in Section~\ref{sec:Defense}, model training plays a key role in achieving robustness against adversarial attacks. 
The issue of distinguishing the appearance of point clouds from one class versus another class may present significant challenges. For example, a ``flower pot'' looks similar to a ``cup'' (see Figure~\ref{fig_saliency}) due to its conical shape, so it does not take much to make a deep model misclassify one for another. From this point of view, models should be trained to be very strong at distinguishing classes whose appearance is similar. This would help improve not only robustness against attacks but also the overall accuracy and generalizability.

A basic premise in statistical ML~\cite{abu-mostafa2012LFD} is that simpler models generalize better, although they may not be as accurate as more complex models. Since adversarial attacks often involve perturbations of the original point cloud, this would seem to imply that simpler models are less likely to be fooled by them. From this point of view, the choice of a model is a trade-off between accuracy, which generally requires higher complexity, and robustness (to attacks, as well as unseen data), which seems to favor not-too-high complexity. Since accuracy is the predominant factor of usefulness of a model, the trend has been towards more complex models, but with additional regularization~\cite{abu-mostafa2012LFD} and more sophisticated learning strategies to strengthen the robustness.

\subsection{Future directions}

\subsubsection{Transferability}
In the context of adversarial attacks, the term \textbf{transferability} refers to the ability of an attack against a given target model to be effective against a different, potentially unknown model. Transferable attacks are not tied the specifics of any one model, but target more fundamental issues, and are therefore also useful in broadening the understanding of adversarial attack and defense principles. Currently, there is a limited amount of research on transferable attacks on 3D point clouds ~\cite{hamdi2020advpc,liu2022boosting,he2023generating,liu2022imperceptible}, so this is one potentially fruitful direction for future research. 

\subsubsection{New tasks} 
Presently, most adversarial attack research is focused on the classification task. This was in part influenced by the wide availability of datasets and related classification models. However, in practice, the role of adversarial attacks is to disrupt a complex system, which may involve other tasks such as detection, segmentation, tracking, etc. It is important to study adversarial attacks and defenses in these more general settings, 
in order to gain a comprehensive understanding of their performance in diverse applications.

\subsubsection{Point cloud attributes} 
The vast majority of adversarial attacks and defenses related to point clouds have focused on point-cloud geometry. However, point clouds may also have attributes such as color~\cite{8i2017color}. Changing the color of points in a point cloud may disrupt classification, segmentation, and other analysis tasks, hence attributes are a potential target for attacks. Since the color attributes of a point cloud play a similar role to the pixel colors in an image, 2D attacks and defenses may provide useful guidelines for initiating the work in this area. Moreover, this would open up possibilities for creating attacks and defenses that simultaneously consider geometry and attributes, a previously unexplored topic.

\section{Conclusion}
\label{sec:Conclusion}

Adversarial attacks on 3D point cloud classification have become a significant concern in recent years. These attacks are able to manipulate 3D point clouds in a way that leads the victim model(s) to make incorrect decisions with potentially harmful consequences. Adversarial attacks on 3D point clouds can be categorized according to the methodologies employed to modify the point cloud, and may have additional attributes in terms of the location of the attack, target type, and adversarial knowledge. We have reviewed a variety of attack methodologies, with examples from the existing literature, highlighting their main characteristics and their relationships. 

To defend against these attacks, researchers have proposed two main categories of approaches: data-focused and model-focused. Data-focused techniques attempt to undo adversarial modifications on the point cloud in order to increase the chance of correct decision, while model-focused approaches attempt to make the model(s) more resilient to adversarial attacks. 
For stronger protection against 
attacks, data-focused and model-focused 
techniques can be combined. 

In addition to reviewing the main attack and defense approaches related to 3D point cloud classification, we also presented the main datasets used in this field, as well as the most widely used victim models. Finally, we summarized the main challenges and outlined possible directions for future research in this field. We hope the article will be helpful to those entering the field of adversarial attacks on 3D point clouds and serve the current research community as a quick reference.

\bibliographystyle{IEEEtranN}
\bibliography{References}

\EOD
\end{document}